\documentclass[12pt,a4paper]{article}

\usepackage[british]{babel}

\usepackage[a4paper,top=2cm,bottom=2cm,left=2.5cm,right=2.5cm,marginparwidth=1.75cm]{geometry}

\usepackage[style=apa, backend=biber]{biblatex}
\DeclareLanguageMapping{british}{british-apa}
\DeclareFieldFormat[article]{volume}{\apanum{#1}}

\usepackage{amsmath}
\usepackage{amsfonts}
\usepackage{mathrsfs}
\usepackage{graphicx}
\usepackage{subcaption}
\usepackage{booktabs}
\usepackage{threeparttable}
\usepackage{diagbox}
\usepackage{url}

\usepackage{algorithm}
\usepackage{algorithmicx}
\usepackage{algpseudocode}
\usepackage{listings}

\usepackage[T1]{fontenc}
\usepackage{csquotes}
\usepackage{lmodern}
\usepackage{helvet}
\usepackage{setspace}
\usepackage[colorlinks=true, allcolors=blue]{hyperref}

\usepackage{titlesec}
\titleformat{\section}
  {\normalfont\Large\bfseries}{\thesection.}{1em}{}

\usepackage{caption}
\date{}

\begin{document}

\begin{center}
{\LARGE Moonworks Lunara Aesthetic I: An Art Dataset} \\[4.2ex]
{\small Yan Wang, Sayeef Abdullah,\\ Partho Hassan, and Sabit Hassan} \\ [2.5em]
{\large Moonworks AI} \\[2.5em]
\end{center}

\begin{figure}[!h]
    \centering
    \setlength{\tabcolsep}{2pt}
    \begin{tabular}{cccccc}
        \includegraphics[width=0.16\textwidth]{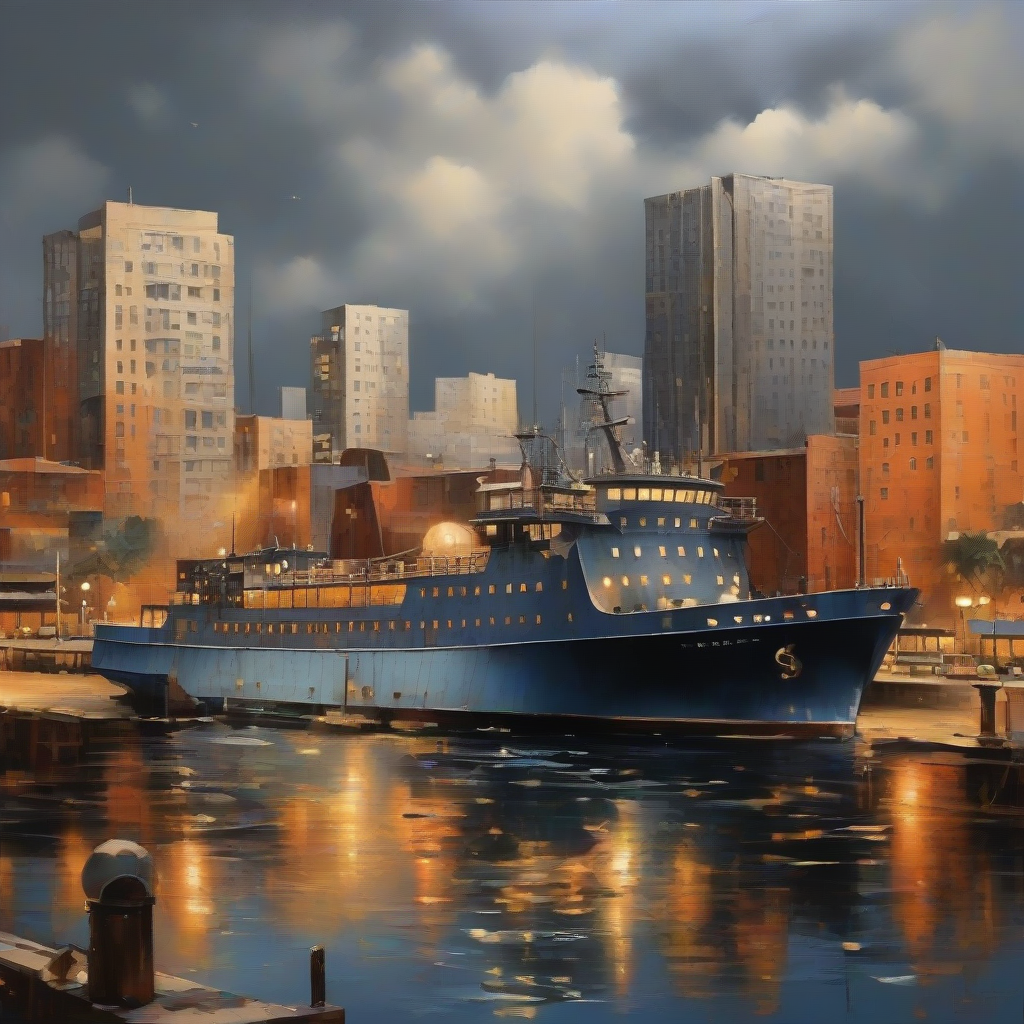} &
        \includegraphics[width=0.16\textwidth]{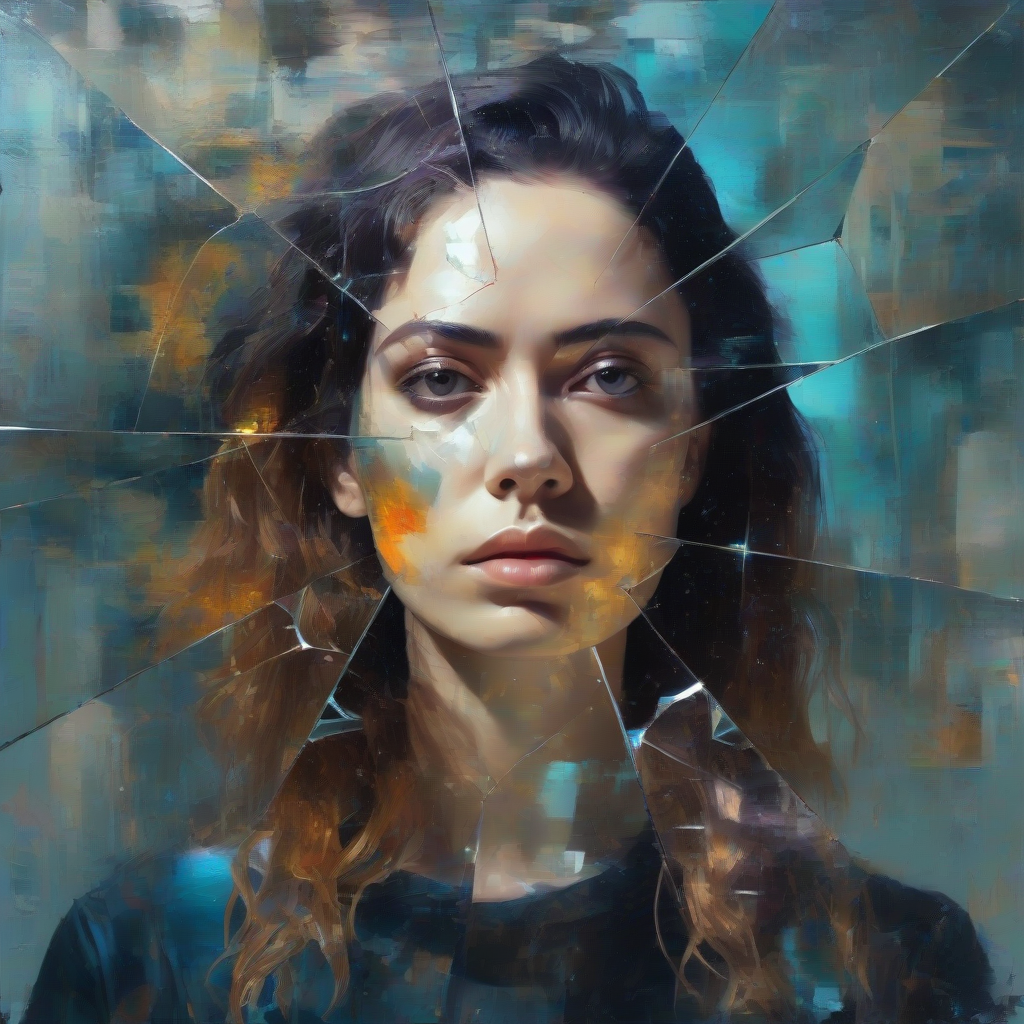} &
        \includegraphics[width=0.32\textwidth]{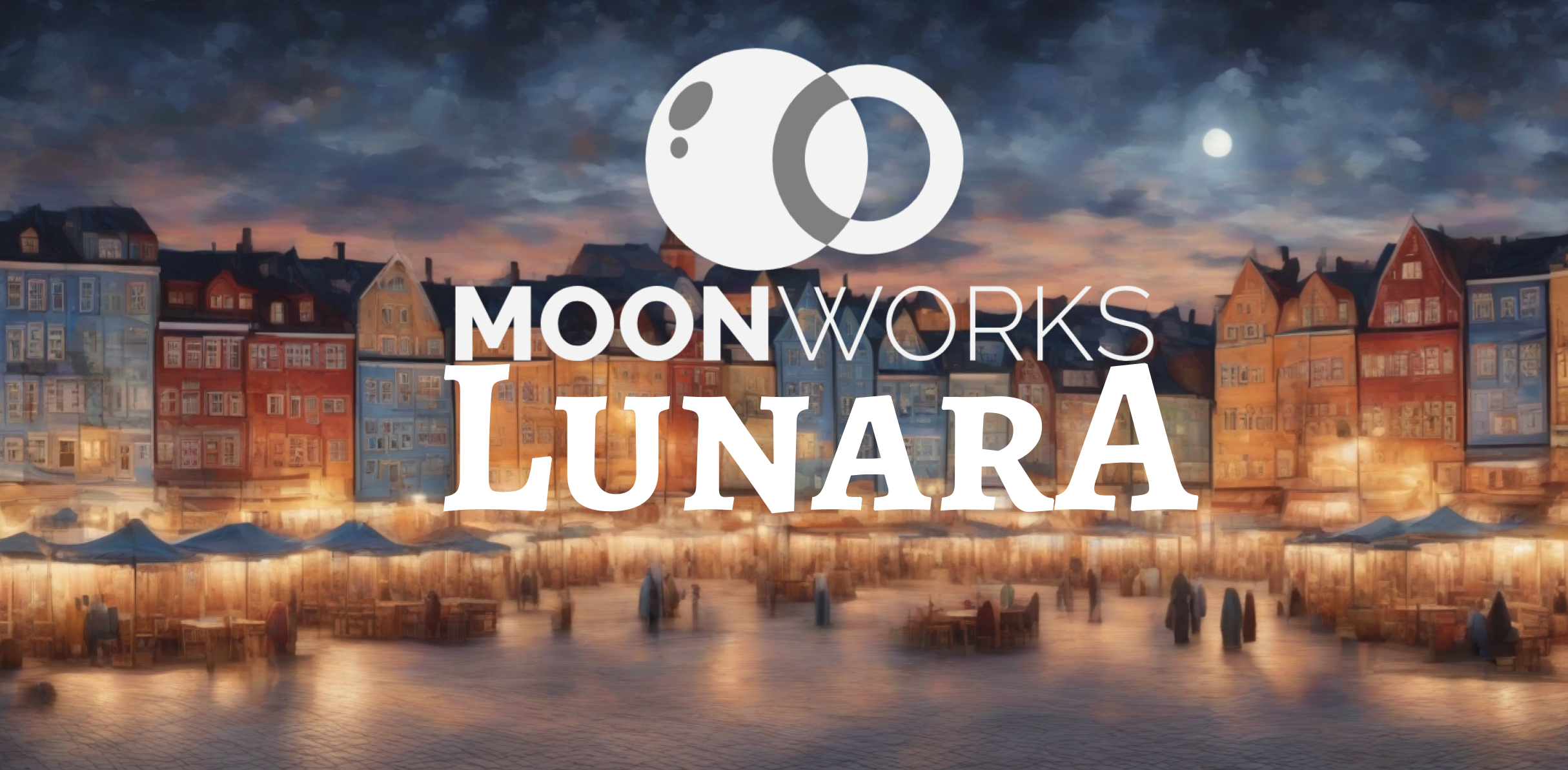}
        \includegraphics[width=0.16\textwidth]{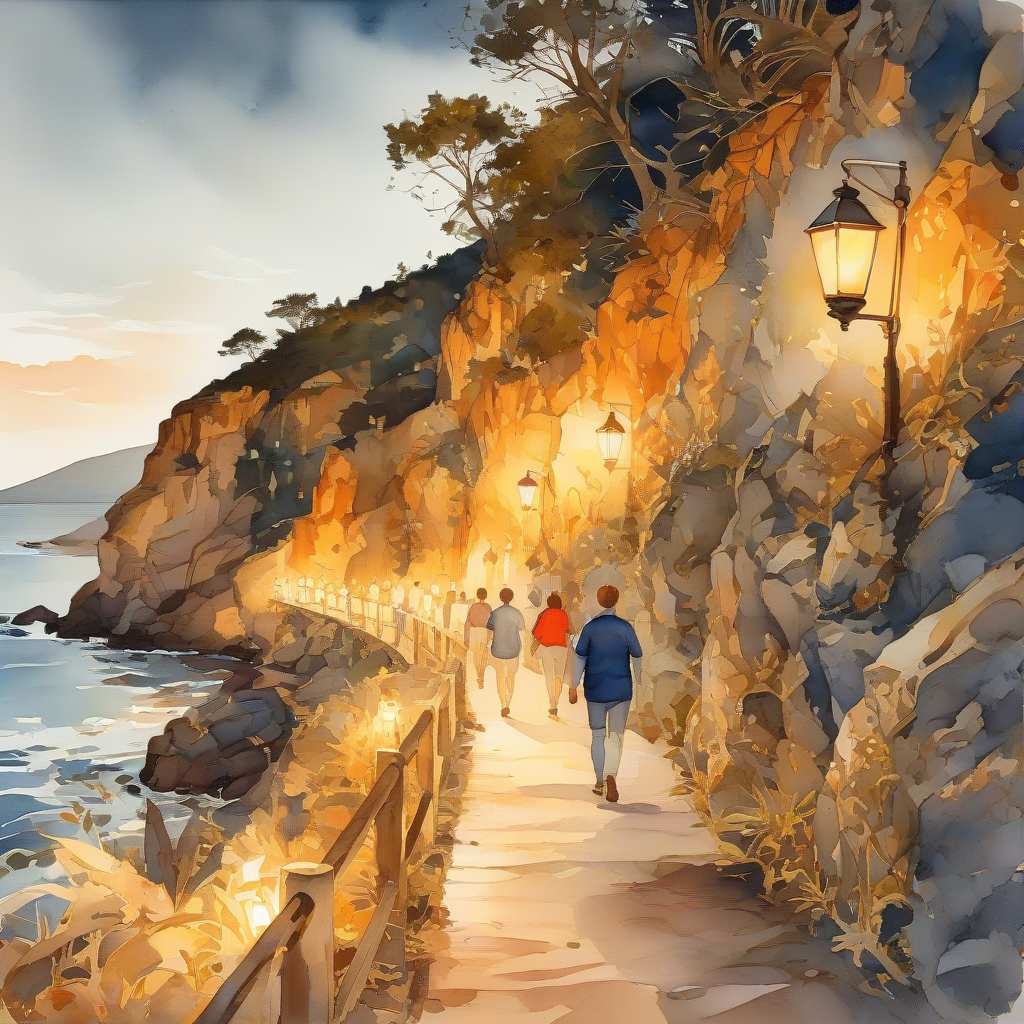} &
        \includegraphics[width=0.16\textwidth]{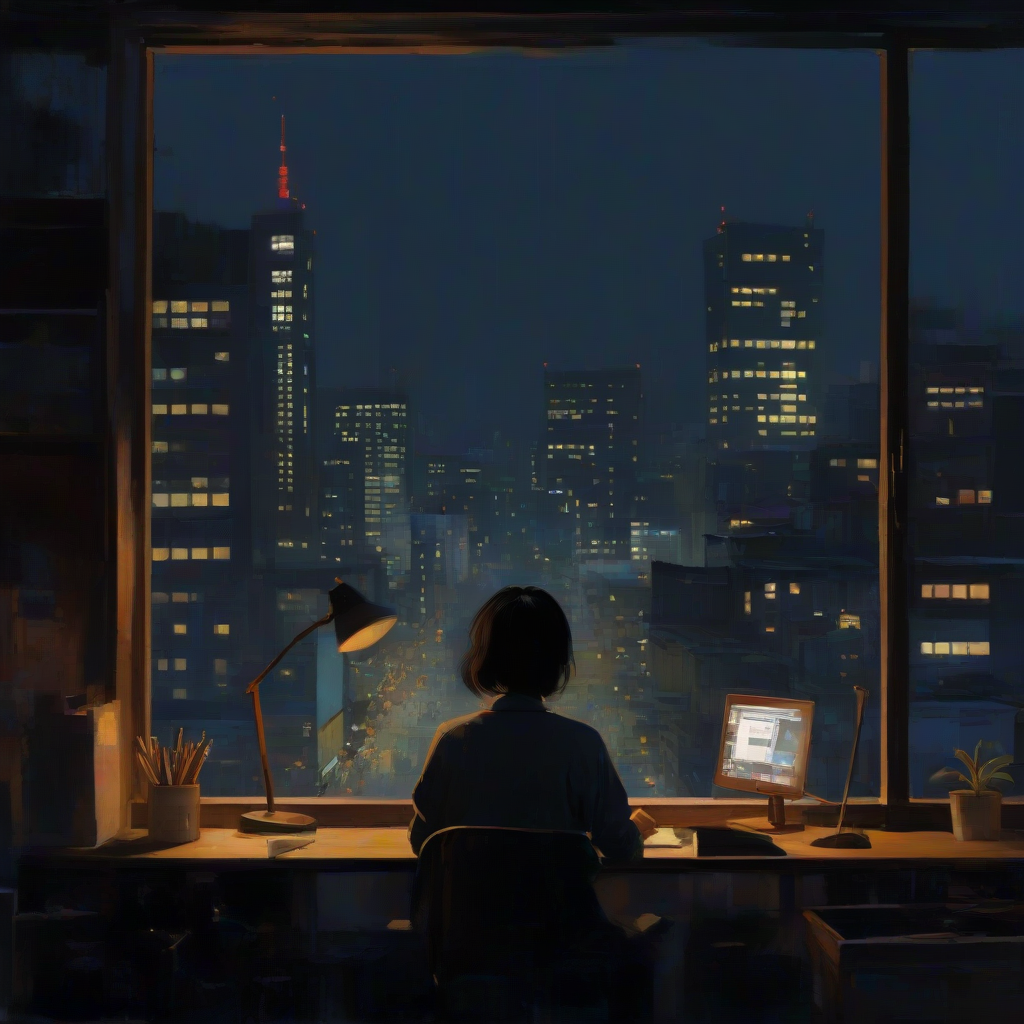}
    \end{tabular}
    \label{fig:style-banner}
\end{figure}

\begin{abstract}
This data card presents the first public release of the Lunara Aesthetic Dataset, a curated set of 2,000 image–prompt pairs for controlled research on prompt grounding and style conditioning in text-to-image generation systems. The dataset spans diverse artistic styles, including regionally grounded aesthetics from the Middle East, Northern Europe, East Asia, and South Asia, alongside general categories such as sketch and oil painting. All images are generated using the Moonworks Lunara model and intentionally crafted to embody distinct, high-quality aesthetic styles, yielding a first-of-its-kind dataset with substantially higher aesthetic scores, exceeding even aesthetics-focused datasets, and general-purpose datasets by a larger margin.  Each image is accompanied by a human-refined prompt and structured annotations that jointly describe salient objects, attributes, relationships, and stylistic cues. Unlike large-scale web-derived datasets that emphasize breadth over precision, the Lunara Aesthetic Dataset prioritizes aesthetic quality, stylistic diversity, and licensing transparency, and is released under the Apache 2.0 license to support research and unrestricted academic and commercial use\footnote{\url{https://huggingface.co/datasets/moonworks/lunara-aesthetic}}.

\end{abstract}

% =========================
% Visual style banner
% =========================

\section{Introduction}
% =========================
% Visual banner (no caption)
% =========================

Recent advances in text-to-image generation have been driven by increasingly capable proprietary and API-served systems, including OpenAI’s native image generation in GPT-4o and the newer ChatGPT Images stack, as well as Google’s Gemini image models (Nano Banana and Nano Banana Pro) (\cite{openai_4o_image_generation,openai_4o_image_system_card_addendum,openai_new_chatgpt_images,google_nanobanana_pro_blog,google_gemini_image_generation_docs}). These systems often deliver strong compositional consistency and high-fidelity rendering, particularly for challenging cases such as legible text and instruction-heavy edits (\cite{openai_4o_image_system_card_addendum,openai_4o_image_generation,google_gemini_image_generation_docs}). However, the outputs of such services are commonly governed by usage terms that restrict using generated content to develop or train competing models, limiting their direct utility as openly reusable training data for competitive model development.

Open-source and publicly reusable alternatives provide greater accessibility, but exhibit complementary limitations. Models such as Stable Diffusion XL (SDXL) enable community-driven development and are widely used as data sources (\cite{podell2023sdxl}), yet generations from smaller open models can still show structural inconsistencies and artefacts in practice under complex compositions, fine-grained spatial constraints, or stylized prompts. At the other end of the spectrum, efficiency-optimized and distilled generators can improve structural integrity and consistency, but are not necessarily specialized for capturing nuanced artistic aesthetics across cultural regions, historical traditions, and diverse media (\cite{zimage2025}).

Beyond model availability, dataset design remains a bottleneck for transparent and reproducible evaluation of prompt-following behavior. Large-scale web-scraped image--text datasets provide broad coverage but frequently pair images with noisy captions or alt-text that differ systematically from instruction-like prompts used in modern text-to-image systems. As a result, failures in prompt adherence are often confounded with annotation noise or underspecification rather than reflecting model behavior. Classic computer vision datasets such as ImageNet (\cite{deng2009imagenet}) are not intended for aesethetic image generation training, while captioning datasets such as CC3M emphasise descriptive language rather than prompt-like instructions (\cite{sharma2018conceptual}). In addition, licensing and provenance in web-scale collections can be difficult to audit, and smaller high-quality datasets used to improve prompt adherence are rarely released publicly.

We introduce the \textit{Lunara Aesthetic Dataset}, a public release of 2,000 image--prompt pairs curated specifically for aesthetic modeling,  style conditioning,  and standardized benchmarking for other models. 

All images are generated by Lunara, a sub-10B param at inference model by Moonworks \footnote{\url{https://moonworks.ai/}} and paired with human-refined prompts that explicitly describe salient objects, attributes, relations, and stylistic cues present in the image, enabling controlled experimentation and reproducible comparison. The dataset spans both modern and traditional artistic styles across four geographical regions---the Nordic region, South Asia, East Asia, and the Middle East, and also includes a region-agnostic set of distinct media-focused categories (e.g., oil painting, sketch, mixed media, and stamp art). 

Lunara was trained on a highly curated dataset and achieves strong aesthetic performance, underscoring the value of curated data construction. Motivated by this result, we release a similarly refined open dataset. We hypothesize that this dataset will achieve higher aesthetic scores than broader general and even datasets curated specifically for aesthetic quality. By making this dataset publicly available, we aim to facilitate reproducible research on aesthetic modeling, data quality, and controlled style learning in modern image generation systems.

\section{Dataset Overview}

\begin{table}[]
\small
\centering
\label{tab:dataset_and_pos}
\begin{minipage}[t]{0.48\linewidth}
\centering
\begin{tabular}{l c}
\hline
\textbf{Statistic} & \textbf{Value} \\
\hline
Total images & 2,000 \\
Topics|Regions|Styles & 7|5|17 \\
Avg. prompt length (tokens) & 18.3 \\
Avg. prompt length (chars) & 130.8 \\
Image resolution & $1024 \times 1024$ \\
\hline
 & Serene, misty,\\ 
Top keywords & cinematic, lighting\\
& dreamy, glow\\
\hline
\end{tabular}
\end{minipage}
\hfill
\begin{minipage}[t]{0.48\linewidth}
\small
\centering
\begin{tabular}{l r}
\hline
\textbf{POS Tag} & \textbf{Count} \\
\hline
Noun (NOUN) & 14.7K \\
Adjective (ADJ) & 10.4K \\
Adposition (ADP) & 2.1K \\
Proper Noun (PROPN) & 8.0K \\
Verb (VERB) & 1K \\
Determiner (DET) & 0.1K \\
Other & 0.28K \\
\hline
Total & 36.6K \\
\hline
\end{tabular}

\end{minipage}
\caption{Dataset statistics and prompt analysis.}

\end{table}

The \textbf{Lunara Art Dataset} consists of \textbf{2,000 images} curated to cover a diverse range of visual themes, cultural regions, and artistic styles. All images are provided at a uniform resolution of 1024 $\times$ 1024 pixels, ensuring consistency across the dataset.

\subsection{Prompt Characteristics}
Each image is paired with a descriptive text prompt. The prompts have an average length of 18.3 tokens (approximately 130.8 characters), balancing semantic richness with conciseness. Keyword frequency analysis indicates a strong stylistic emphasis on atmospheric descriptors, with commonly occurring terms such as \emph{serene}, \emph{misty}, \emph{cinematic}, \emph{lighting}, \emph{dreamy}, and \emph{glow}.

A part-of-speech (POS) analysis over 36.6K tokens shows that prompts are dominated by nouns (14.7K) and adjectives (10.4K), reflecting a focus on visual entities and aesthetic attributes. Proper nouns (8.0K) frequently reference specific locations or cultural contexts, while verbs (1.0K), adpositions (2.1K), and determiners (0.1K) occur less frequently.

\subsection{Topic Distribution}
The dataset spans seven high-level topics. Nature \& Landscape constitutes the largest portion (607 images), followed by Everyday Life (503) and Portraits \& Human Figures (346). Additional categories include City, Building \& Architecture (222), Rural \& Agrarian Life (135), Work, Hobby \& Occupations (98), and Religion, Spirituality \& Ascetic Life (89). This distribution supports both scene-centric and human-centric visual understanding tasks.

\begin{figure}[t]
    \centering
    \includegraphics[width=0.7\textwidth]{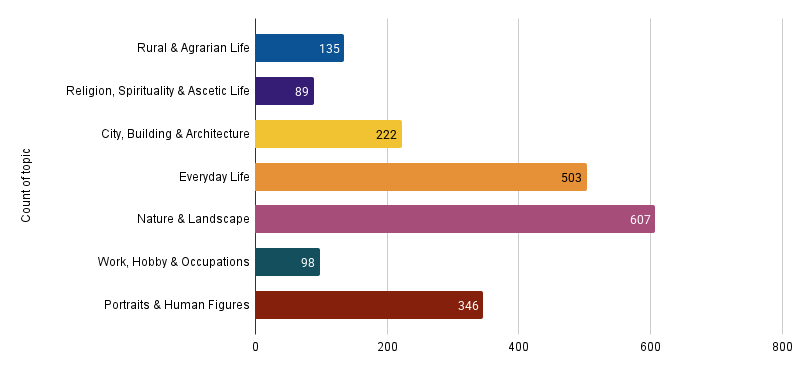}
    \caption{Distribution of key topics in the Lunara Aesthetic Dataset.}
    \label{fig:topic-dist}
\end{figure}

\begin{figure}[t]
    \centering
    \includegraphics[width=0.9\textwidth]{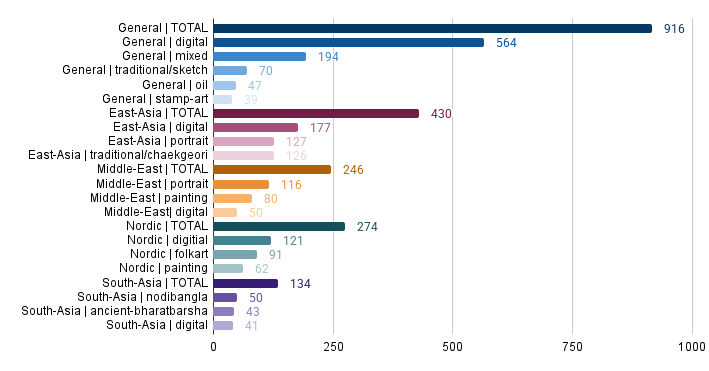}
    \caption{Style distribution of the Lunara Aesthetic Dataset across four geographical regions as well as region-agnostic category.}
    \label{fig:art-dist}
\end{figure}

\subsection{Regional and Style Distribution}
The dataset includes 17 region--style combinations grouped into broader cultural regions. The General category is the most prevalent (916 images), primarily composed of digital art (564) and mixed media (194), with smaller subsets of traditional sketch, oil, and stamp-art styles.

Region-specific subsets include East Asia (430 images), with strong representation of digital, portrait, and traditional \emph{chaekgeori}-inspired styles; Nordic (274), featuring digital, folk art, and painting; and the Middle East (246), balancing portrait, painting, and digital works. South Asia (134) contributes a mix of digital art, \emph{noodibangla}, and ancient \emph{Bharatvarsha}-inspired styles.

Overall, the Lunara Art Dataset provides a balanced combination of thematic breadth, cultural diversity, and stylistic variation, making it suitable for research in text-to-image generation, style analysis, and cross-cultural visual representation.

\subsection{Intended Use}

This dataset is intended to support fine-tuning and adaptation experiments in which interpretability, controllability, and reproducibility are critical. For example, researchers may use the dataset to finetune image generation models to acquire specific aesthetic, regional, cultural, and medium-specific style conditioning, enabling systematic analysis of stylistic learning. 

In addition, the dataset can serve as a standardized benchmark for evaluating the aesthetic quality of generated images and for conducting comparative analyses against Lunara and other image generation models. 

Beyond generation, the dataset's high-quality, human-curated annotations make it suitable for improving topic understanding in VLMs. Finally, the dataset can be used for image retrieval tasks, allowing researchers to identify and study images that are stylistically or semantically similar to those contained in the dataset  

The dataset is not intended to represent the full diversity of real-world imagery or natural language usage.

\section{Dataset Creation and Annotation Pipeline}
Images in this dataset are generated using Moonworks Lunara, a sub-10B parameter model at inference with a proprietary diffusion mixture with transformer block architecture. Lunara is trained using Moonworks CAT (Composite Active Transfer) method, which draws on the literature of active learning~(\cite{hassan2025coherence,hassan-etal-2025-active,hassan-etal-2024-active,hassan-alikhani-2023-calm,Hassan2018interactive}). Active learning iteratively trains models on selective data points; as iterations grow, the training set evolves and focuses on improving model behavior through targeted additions rather than brute-force scale. CAT is the first to formulate and apply active learning for image generation tasks as well as a diffusion mixture architecture. Lunara is trained with a combination of proprietary art and photography datasets constructed in collaboration with affiliated artists and photographers, their semantic variations, as well as public domain data (CC by 4.0) and open-sourced datasets and synthetic data with permissive license (Apache 2.0/MiT) \footnote{further details of Lunara to follow in subsequent releases}. 

\begin{figure}[t]
    \centering
    \includegraphics[width=\textwidth]{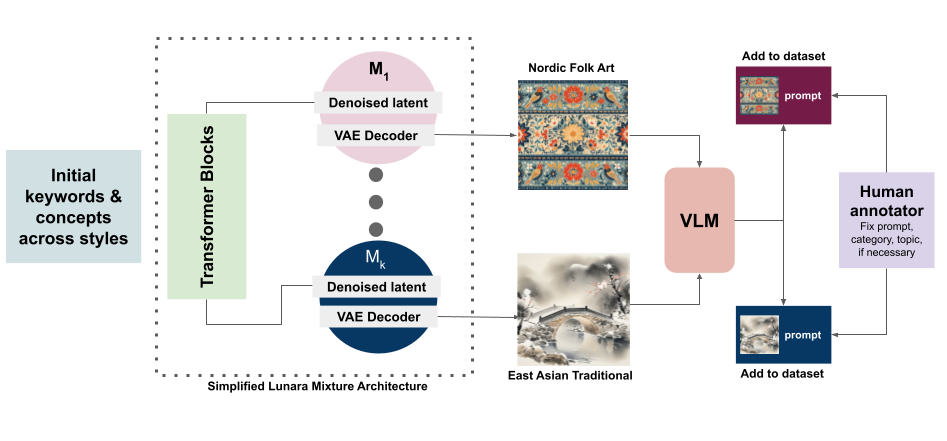}
    \caption{Overview of the Lunara image generation and annotation pipeline, illustrating model-based generation followed by human prompt refinement and filtering.}
    \label{fig:generation-pipeline}
\end{figure}

For each generation step, a subset of the mixture is activated, specializing in a particular regional and artistic style. After generation, human annotators review and refine the prompts to ensure semantic correctness, clarity, and completeness. This refinement process includes correcting factual inaccuracies, resolving ambiguities, and removing descriptions that are not supported by the corresponding image. Image–prompt pairs that do not meet these quality criteria are excluded from the final release.

In addition, each image–prompt pair is annotated with topical labels. Topic and artistic category annotations are obtained through a two-round annotation process. In the first round, annotators propose stylistic, regional, and topical labels from a broad candidate set. Prior to the second round, these labels are consolidated into seven high-level topics, along with the region and style categories shown in Figures~\ref{fig:topic-dist} and~\ref{fig:art-dist}. All images are then reannotated using this unified taxonomy. The two-stage annotation process helps ensure consistency, precision, and high annotation quality across the dataset.

% \section{Qualitative Style Conditioning Examples}

% =========================
% Figure: Regional styles
% =========================
\begin{figure}[t]
    \centering

    % -------- Row 1 --------
    \begin{subfigure}[t]{0.23\textwidth}
        \centering
        {\tiny East Asia Digital}\par
        \includegraphics[width=\linewidth]{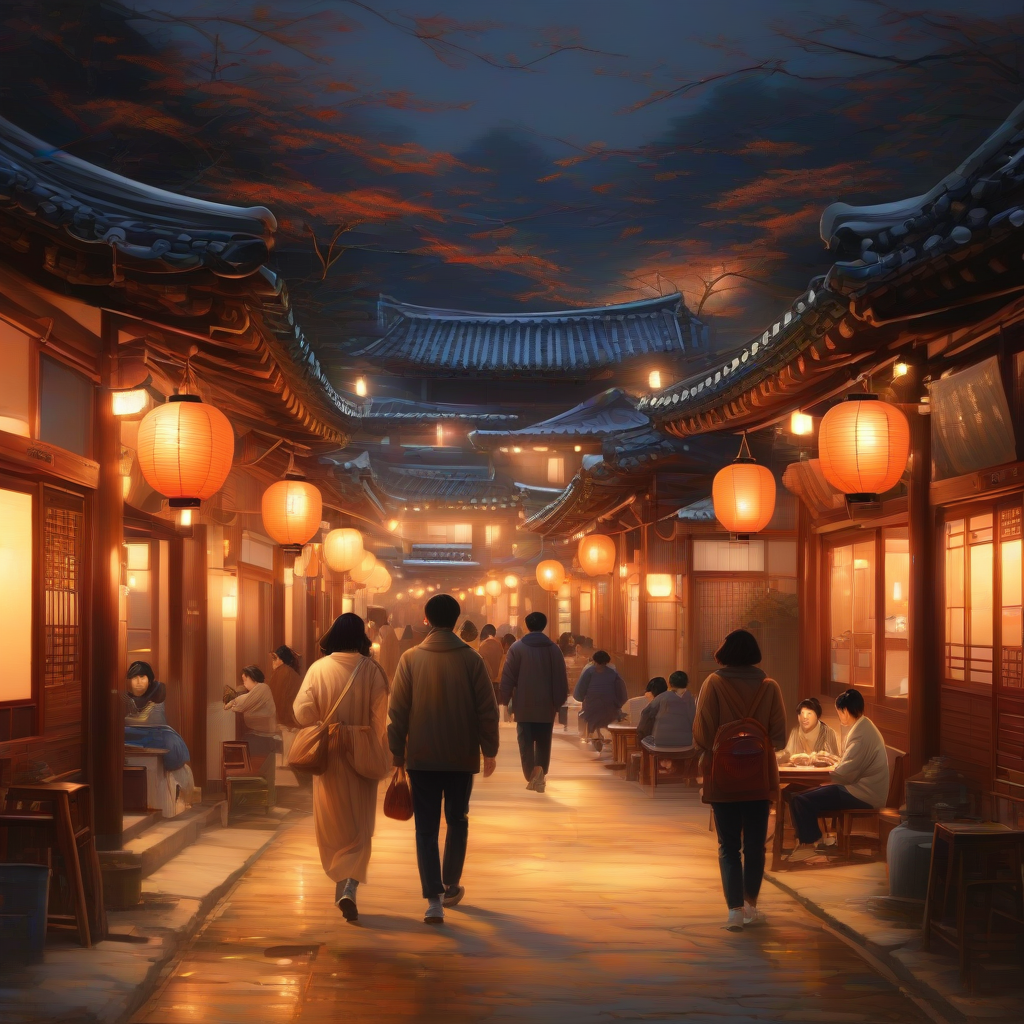}\par
        {\tiny\parbox{\linewidth}{\centering Warm lantern-lit traditional alley, cozy twilight, people strolling, soft golden glow.}}
    \end{subfigure}\hfill
    \begin{subfigure}[t]{0.23\textwidth}
        \centering
        {\tiny East Asia Traditional}\par
        \includegraphics[width=\linewidth]{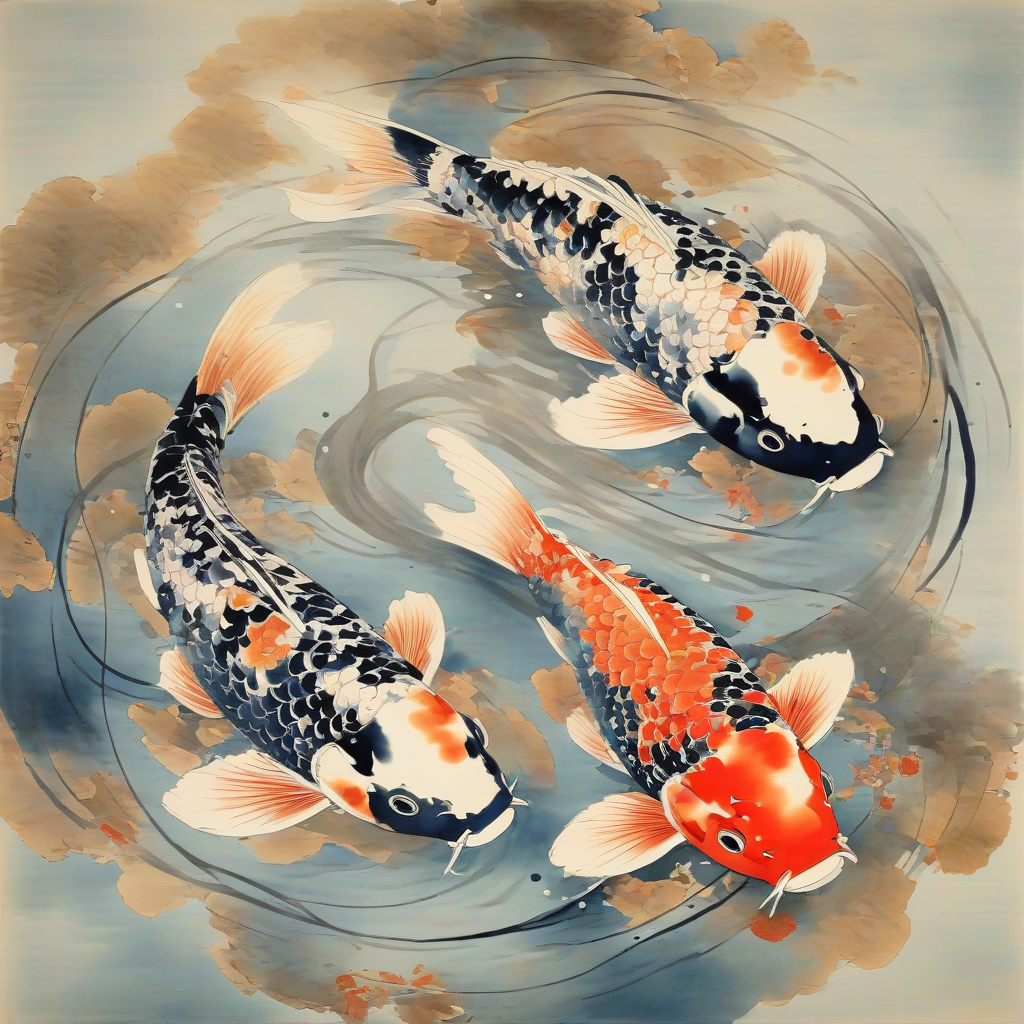}\par
        {\tiny\parbox{\linewidth}{\centering Three koi fish swirling in serene pond, soft light, watercolor style.}}
    \end{subfigure}\hfill
    \begin{subfigure}[t]{0.23\textwidth}
        \centering
        {\tiny East Asia Portrait}\par
        \includegraphics[width=\linewidth]{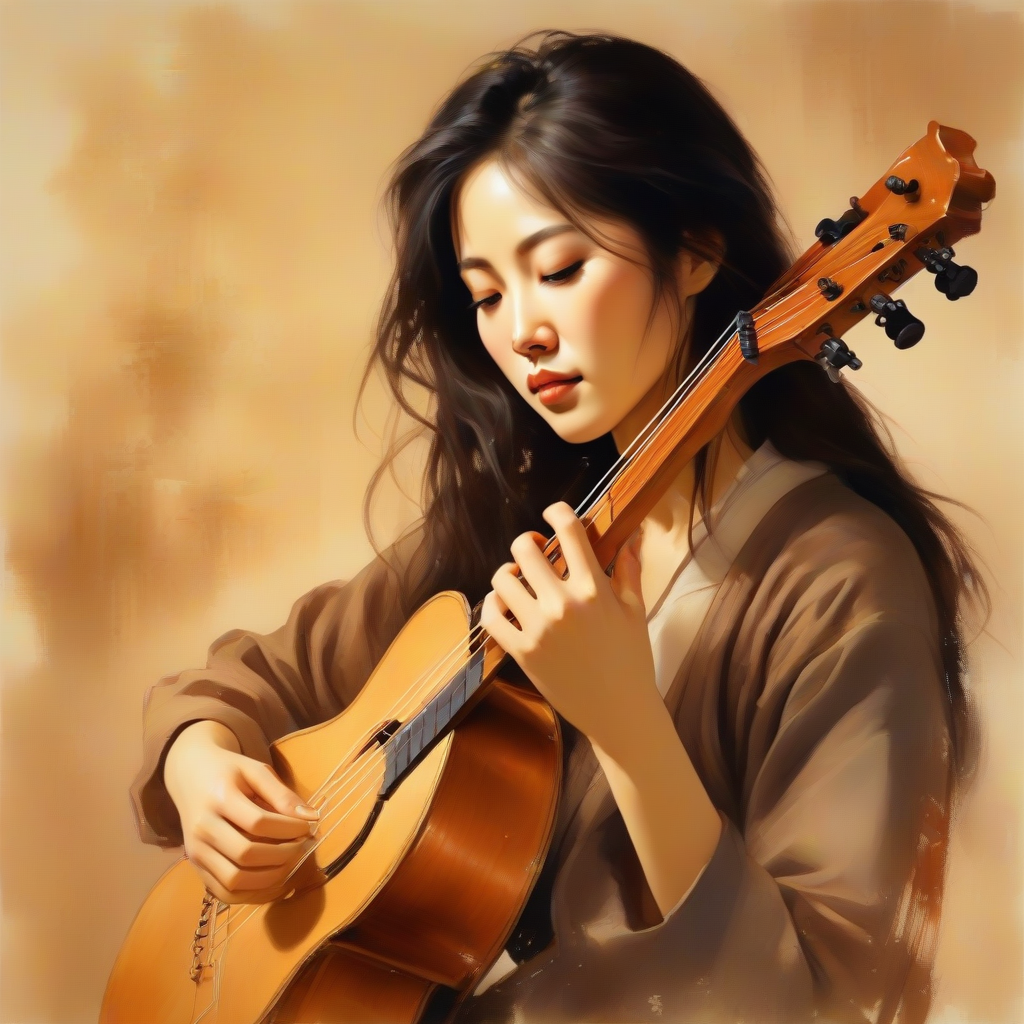}\par
        {\tiny\parbox{\linewidth}{\centering A serene woman playing acoustic guitar, warm lighting, soft bokeh background}}
    \end{subfigure}\hfill
    \begin{subfigure}[t]{0.23\textwidth}
        \centering
        {\tiny Middle East Digital}\par
        \includegraphics[width=\linewidth]{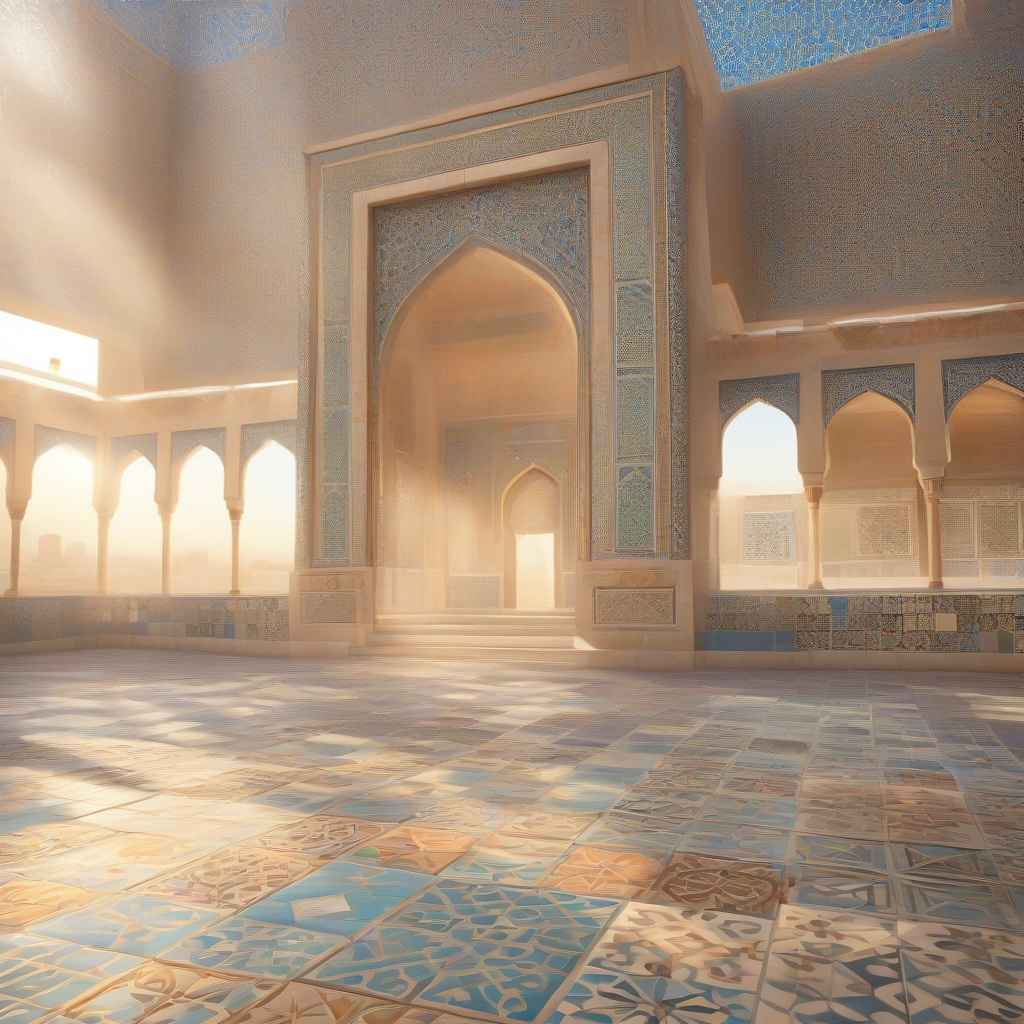}\par
        {\tiny\parbox{\linewidth}{\centering Sunlit Islamic courtyard with intricate blue tiles, warm golden light, wide-angle view.}}
    \end{subfigure}

    \vspace{0.6em}

    % -------- Row 2 --------
    \begin{subfigure}[t]{0.23\textwidth}
        \centering
        {\tiny Middle East Painting}\par
        \includegraphics[width=\linewidth]{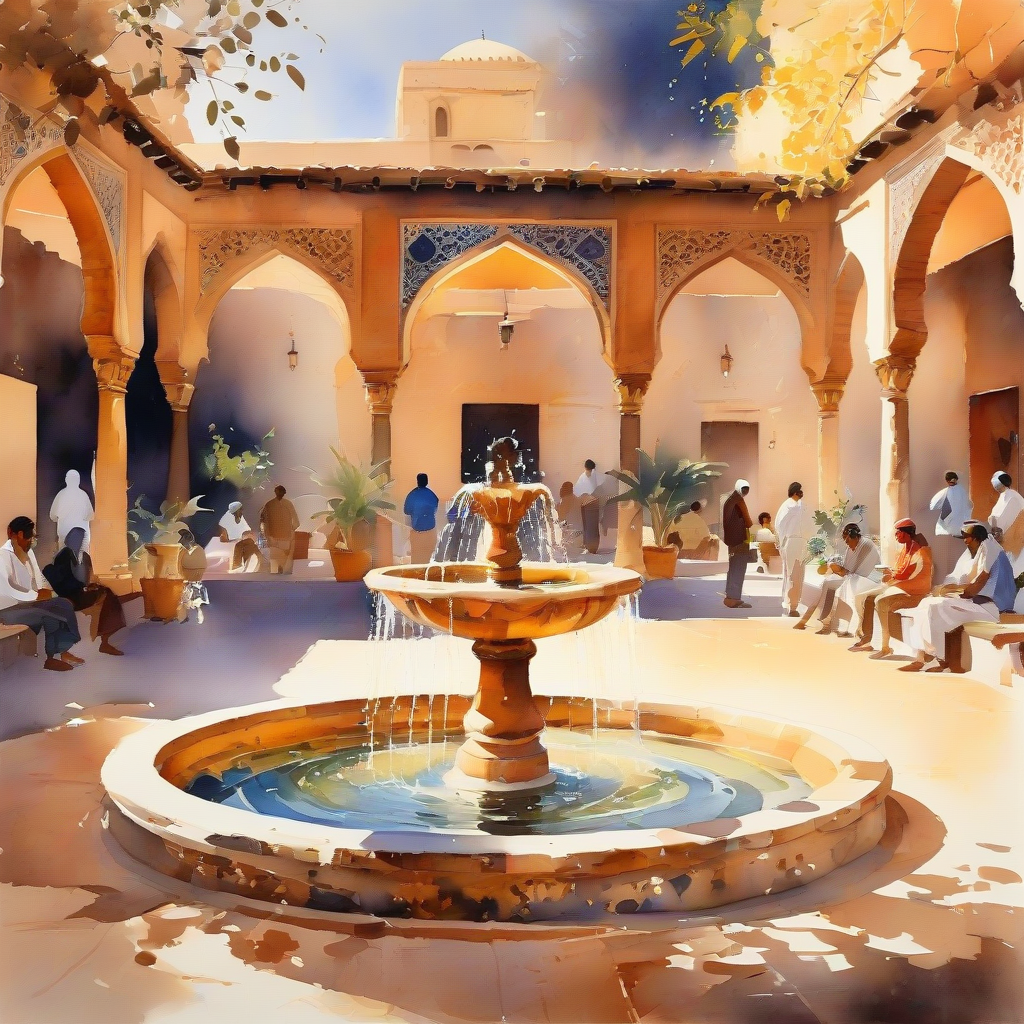}\par
        {\tiny\parbox{\linewidth}{\centering Golden courtyard fountain, warm sunlight, mediterranean arches, serene atmosphere}}
    \end{subfigure}\hfill
    \begin{subfigure}[t]{0.23\textwidth}
        \centering
        {\tiny Middle East Portrait}\par
        \includegraphics[width=\linewidth]{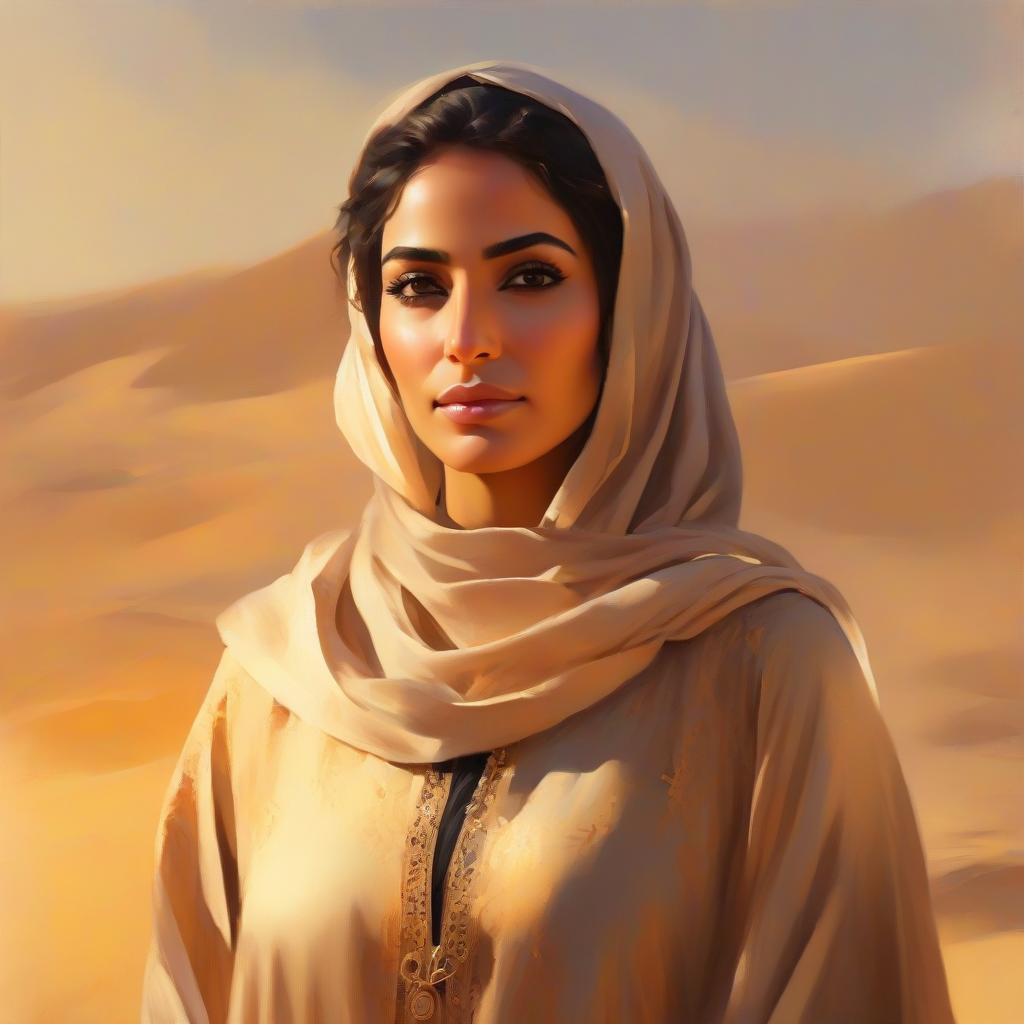}\par
        {\tiny\parbox{\linewidth}{\centering Desert woman in beige hijab, golden light, soft focus, serene expression}}
    \end{subfigure}\hfill
    \begin{subfigure}[t]{0.23\textwidth}
        \centering
        {\tiny Nordic Digital}\par
        \includegraphics[width=\linewidth]{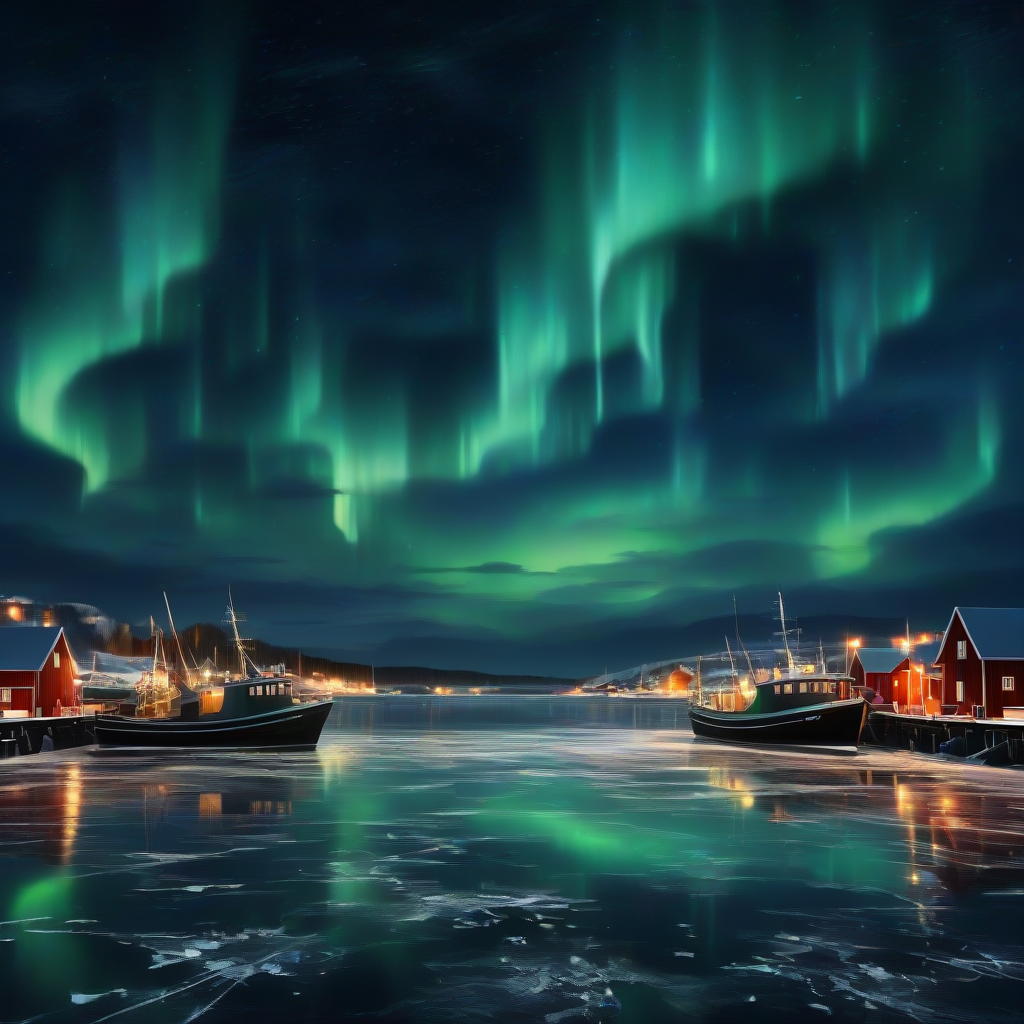}\par
        {\tiny\parbox{\linewidth}{\centering Aurora borealis over icy harbor, red cabins, glowing green lights}}
    \end{subfigure}\hfill
    \begin{subfigure}[t]{0.23\textwidth}
        \centering
        {\tiny Nordic Folkart}\par
        \includegraphics[width=\linewidth]{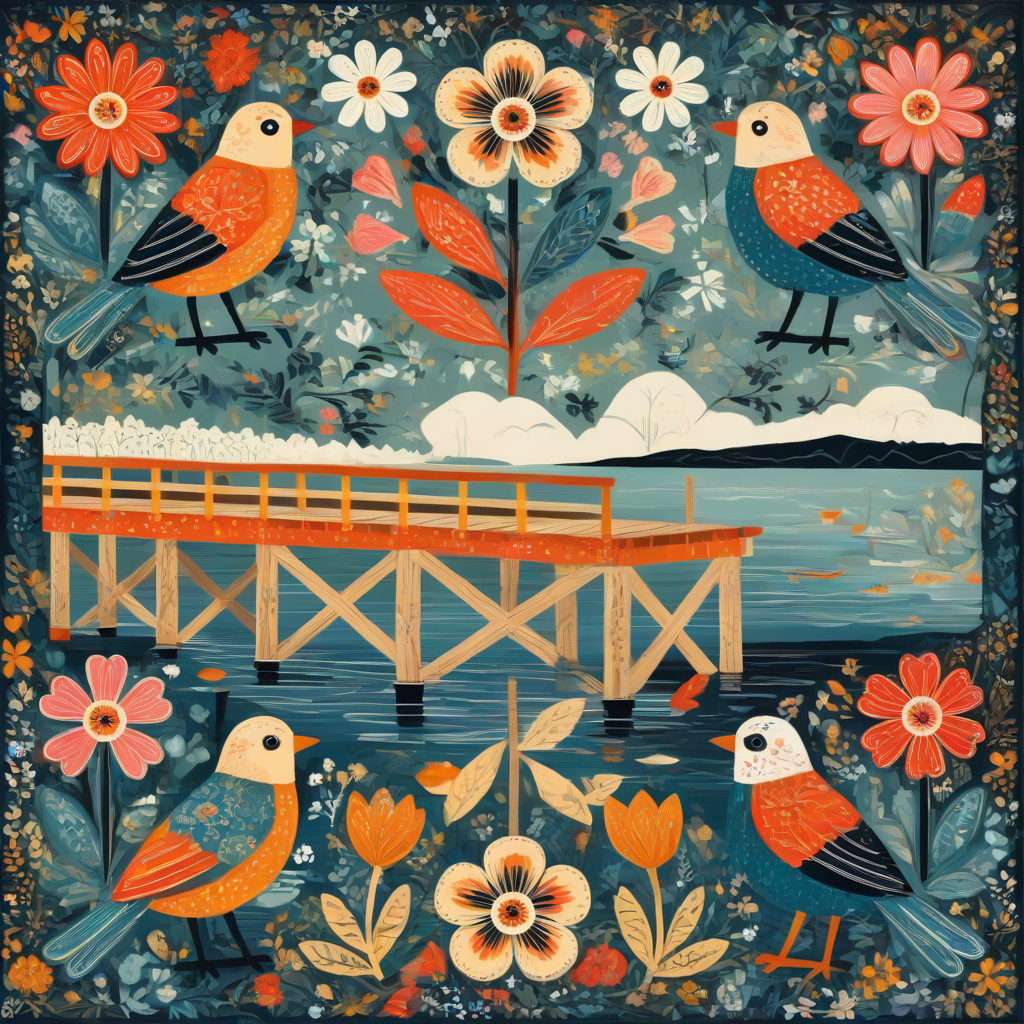}\par
        {\tiny\parbox{\linewidth}{\centering 	Colorful birds on wooden bridge over calm lake, floral background, soft morning light, dreamy art style.}}
    \end{subfigure}

    \vspace{0.6em}

    % -------- Row 3 --------
    \begin{subfigure}[t]{0.23\textwidth}
        \centering
        {\tiny Nordic Aurora}\par
        \includegraphics[width=\linewidth]{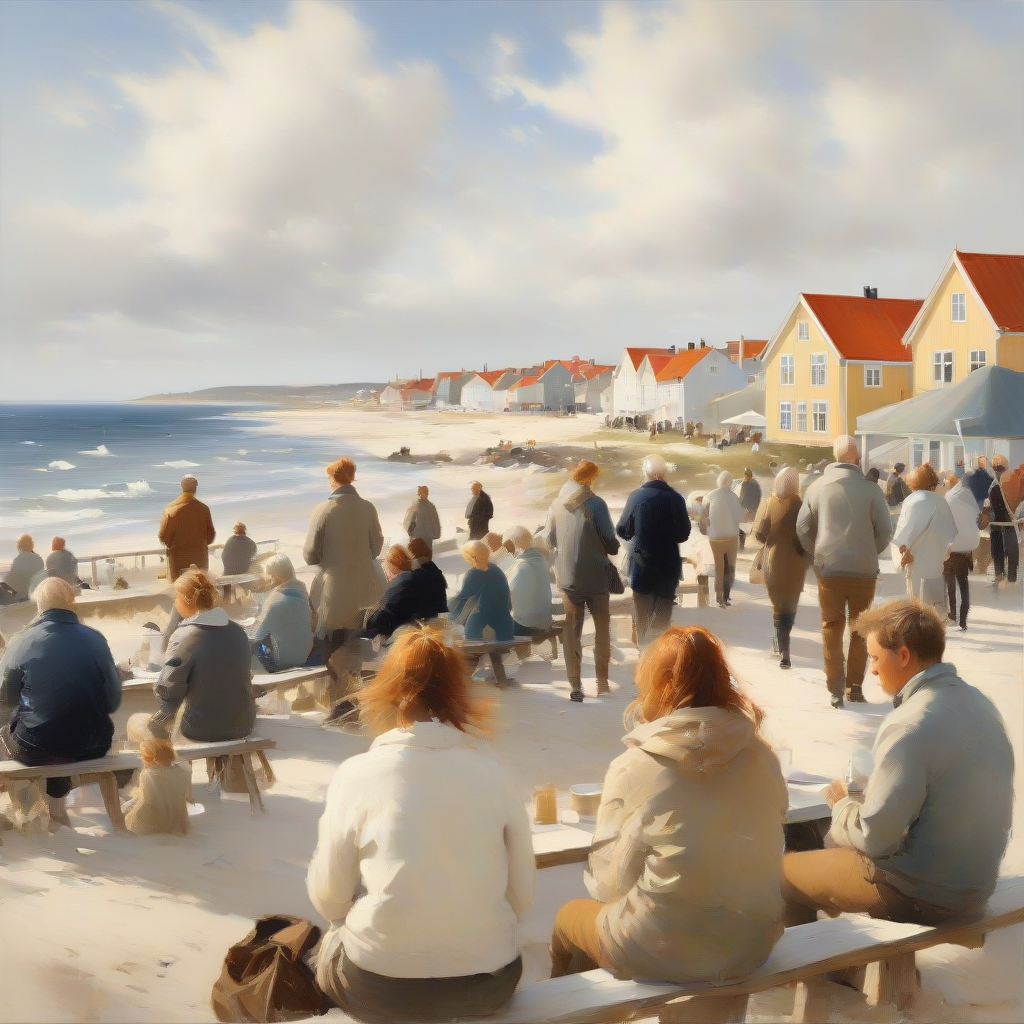}\par
        {\tiny\parbox{\linewidth}{\centering Cozy beachside, soft golden light, people chatting, pastel cottages, gentle waves}}
    \end{subfigure}\hfill
    \begin{subfigure}[t]{0.23\textwidth}
        \centering
        {\tiny South Asia Ancient}\par
        \includegraphics[width=\linewidth]{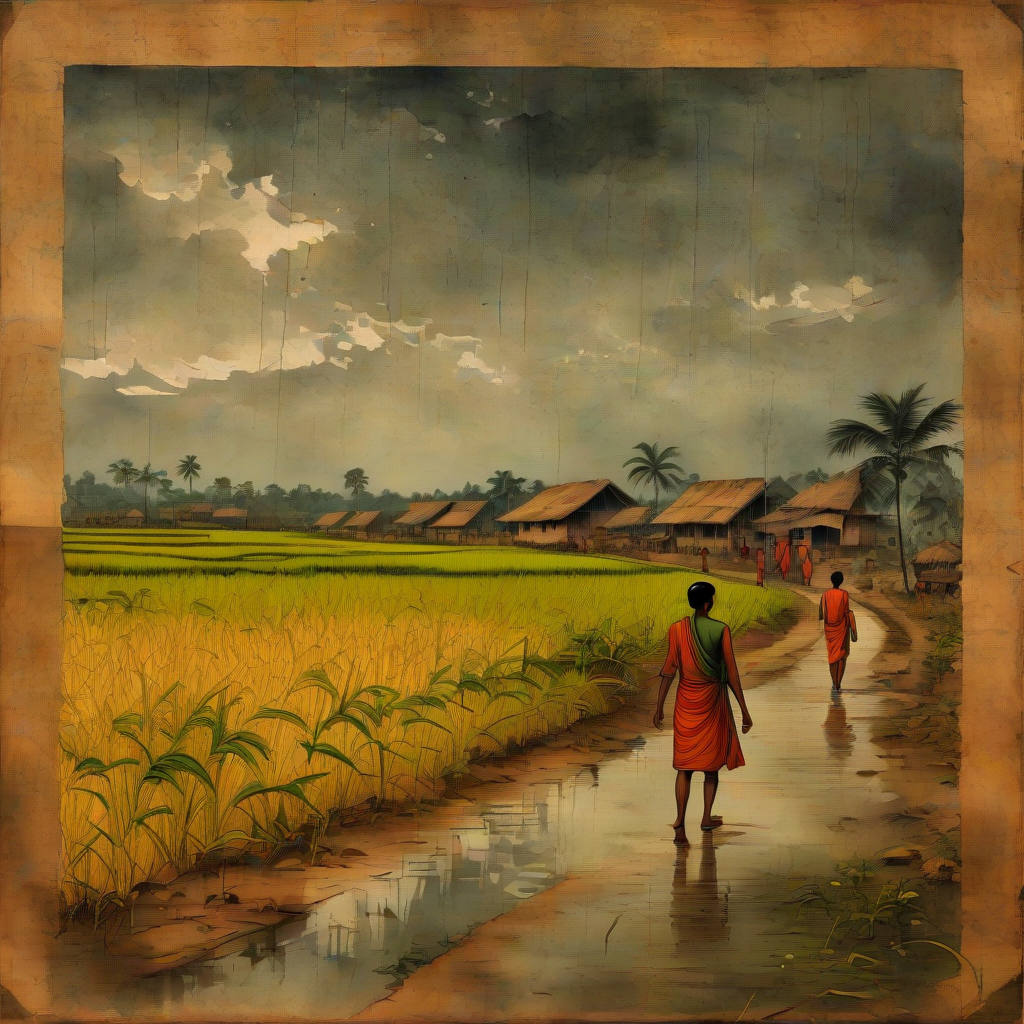}
        {\tiny\parbox{\linewidth}{\centering Rainy village path, vibrant orange robes, golden rice fields, misty sky, vintage painting style}}
    \end{subfigure}\hfill
    \begin{subfigure}[t]{0.23\textwidth}
        \centering
        {\tiny Soth Asia Digital}\par
        \includegraphics[width=\linewidth]{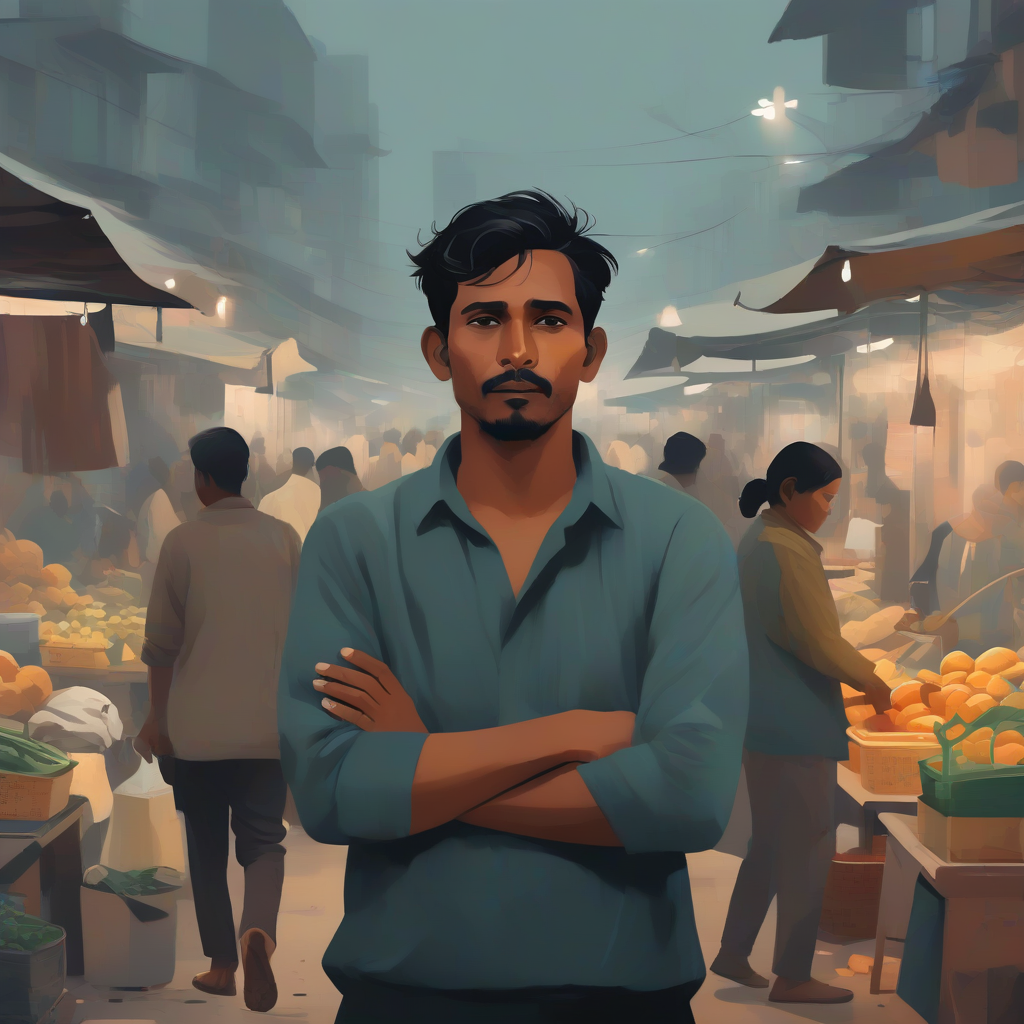}\par
        {\tiny\parbox{\linewidth}{\centering Man in teal shirt stands arms crossed in bustling morning market, soft dawn light}}
    \end{subfigure}\hfill
    \begin{subfigure}[t]{0.23\textwidth}
        \centering
        {\tiny General}\par
        \includegraphics[width=\linewidth]{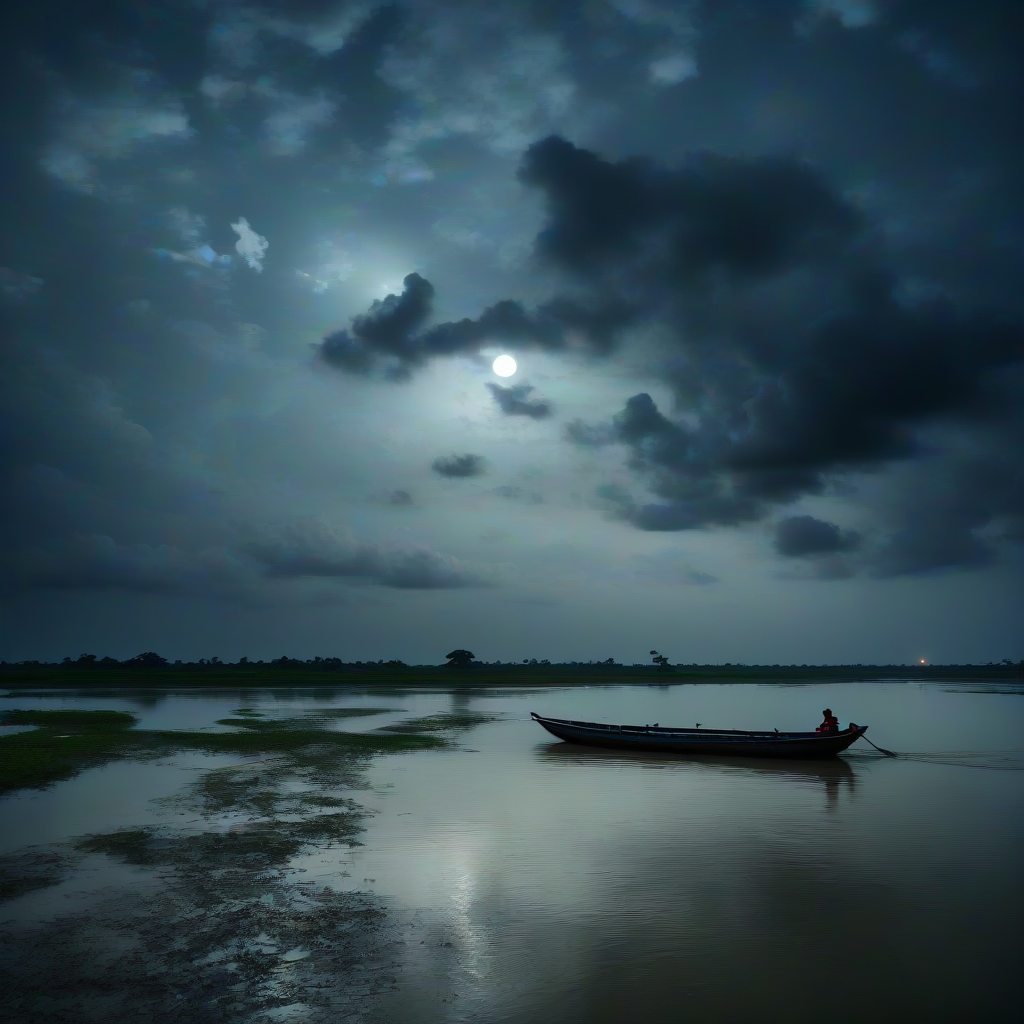}\par
        {\tiny\parbox{\linewidth}{\centering Moonlit river, lone boat, dramatic clouds, moody atmosphere, wide lens, tranquil.}}
    \end{subfigure}

    \caption{Regional style conditioning examples using a shared prompt across multiple cultural aesthetics.}
    \label{fig:regional-styles}
\end{figure}

% =========================
% Figure: General styles
% =========================
\begin{figure}[t]
    \centering

    % -------- Row 1 (3) --------
    \begin{subfigure}[t]{0.3\textwidth}
        \centering
        {\tiny Digital}\par
        \includegraphics[width=\linewidth]{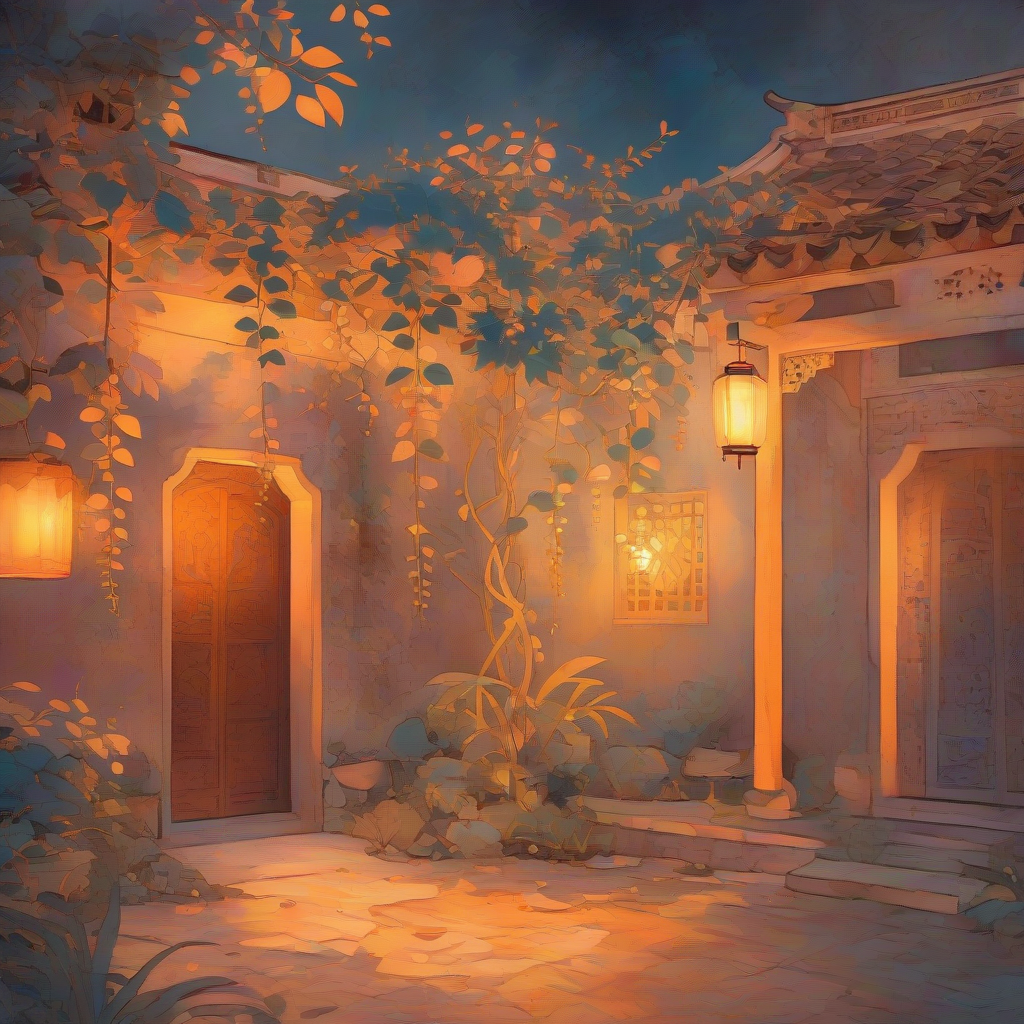}\par
        {\tiny\parbox{\linewidth}{\centering Warm lanterns glow in ancient courtyard, vines cascade, soft twilight ambiance}}
    \end{subfigure}\hfill
        \begin{subfigure}[t]{0.3\textwidth}
        \centering
        {\tiny Digital}\par
        \includegraphics[width=\linewidth]{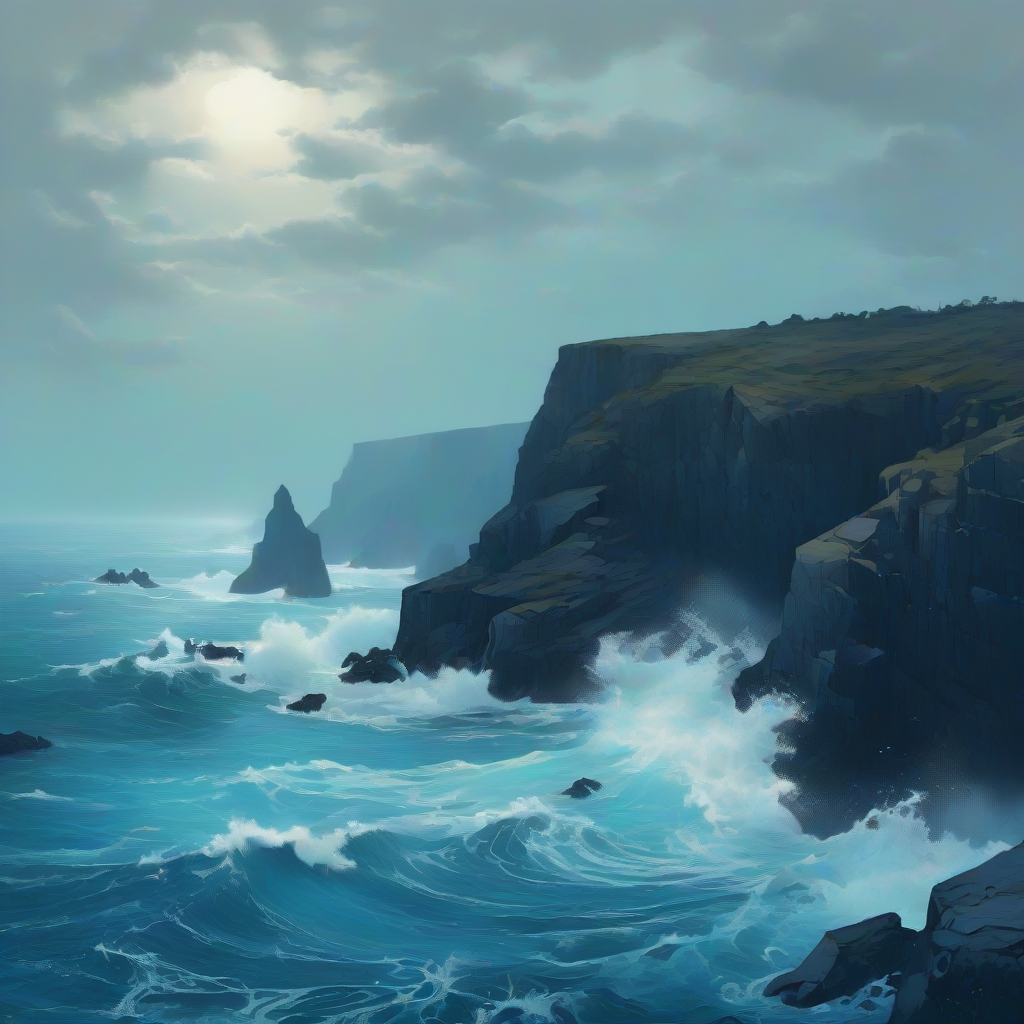}\par
        {\tiny\parbox{\linewidth}{\centering Stormy seas crash against rugged cliffs under moody sky, dramatic lighting, wide-angle view.}}
    \end{subfigure}\hfill
    \begin{subfigure}[t]{0.3\textwidth}
        \centering
        {\tiny Oil}\par
        \includegraphics[width=\linewidth]{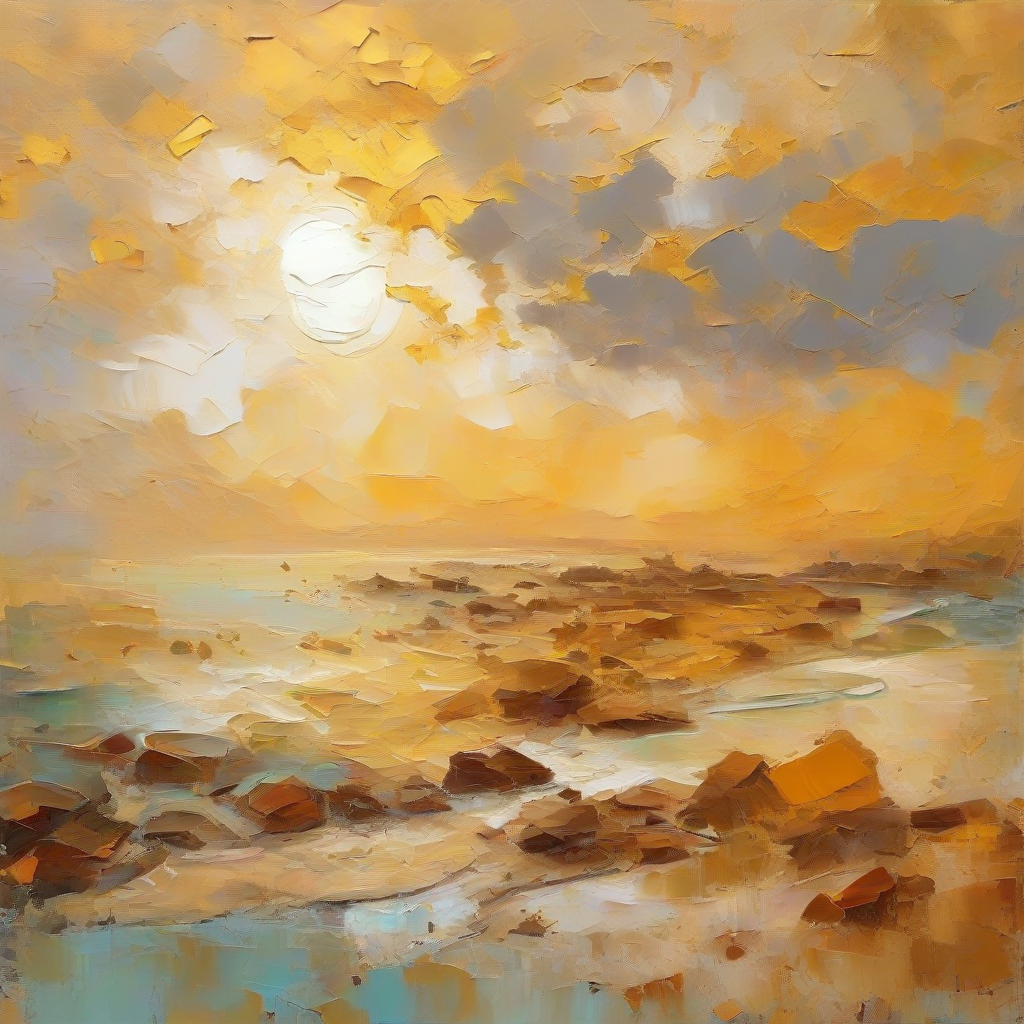}\par
        {\tiny\parbox{\linewidth}{\centering Golden sunset over rocky shore, oil painting, warm tones, wide-angle lens}}
    \end{subfigure}\hfill

    \vspace{0.6em}

    % -------- Row 2 (2) --------
    
    \begin{subfigure}[t]{0.3\textwidth}
        \centering
        {\tiny Sketch}\par
        \includegraphics[width=\linewidth]{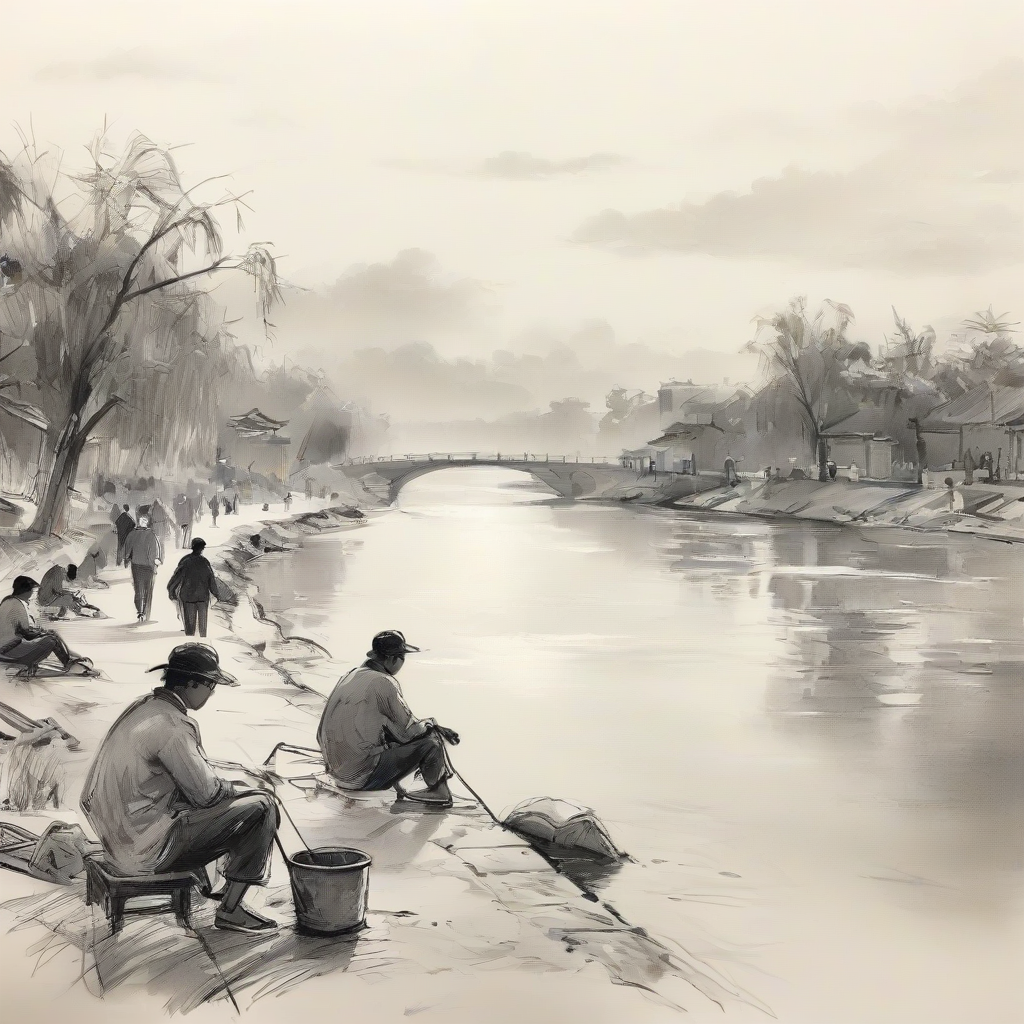}\par
        {\tiny\parbox{\linewidth}{\centering Monochrome riverside fishermen, misty dawn, vintage lens, serene urban landscape}}
    \end{subfigure} \hfill
    \begin{subfigure}[t]{0.3\textwidth}
        \centering
        {\tiny Stamp Art}\par
        \includegraphics[width=\linewidth]{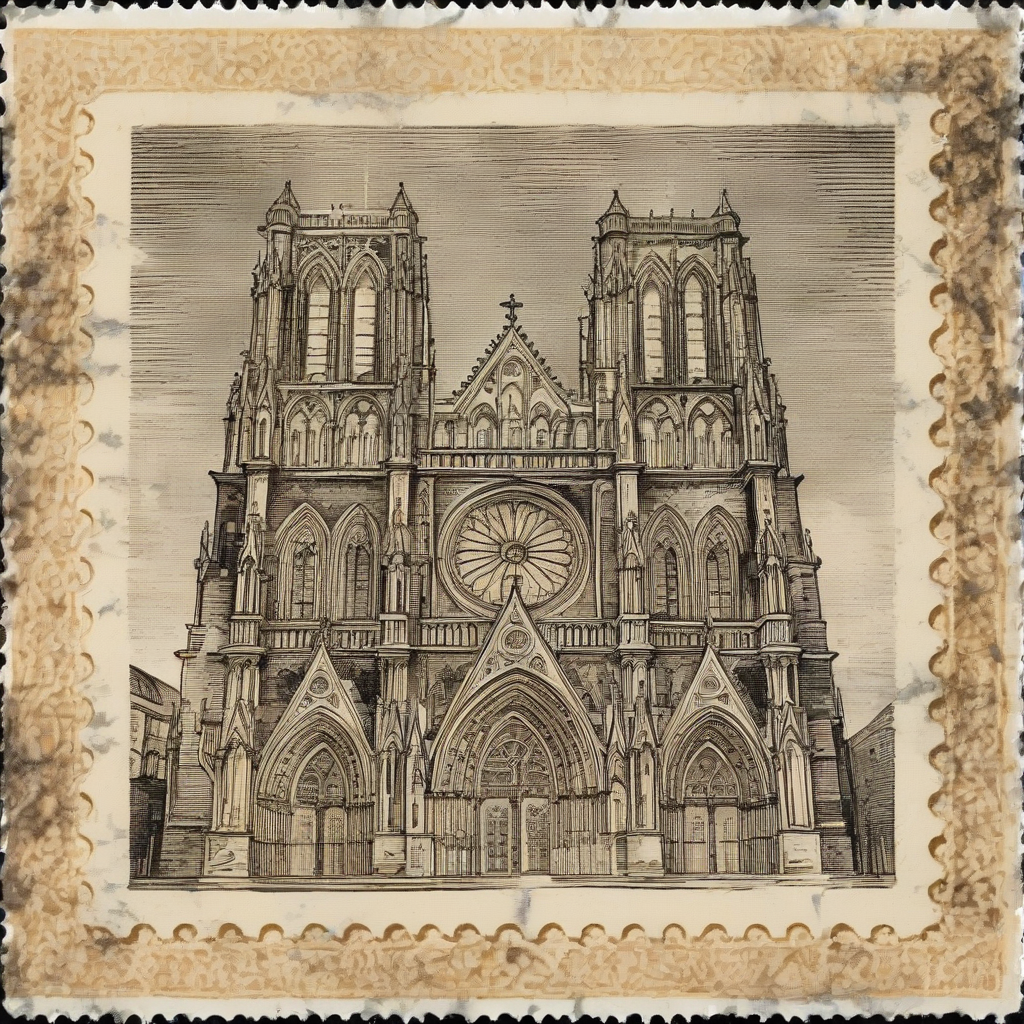}\par
        {\tiny\parbox{\linewidth}{\centering Victorian cathedral, sepia-toned engraving, vintage stamp border, soft focus, historical ambiance}}
    \end{subfigure}\hfill
    \begin{subfigure}[t]{0.3\textwidth}
        \centering
        {\tiny Mixed Art}\par
        \includegraphics[width=\linewidth]{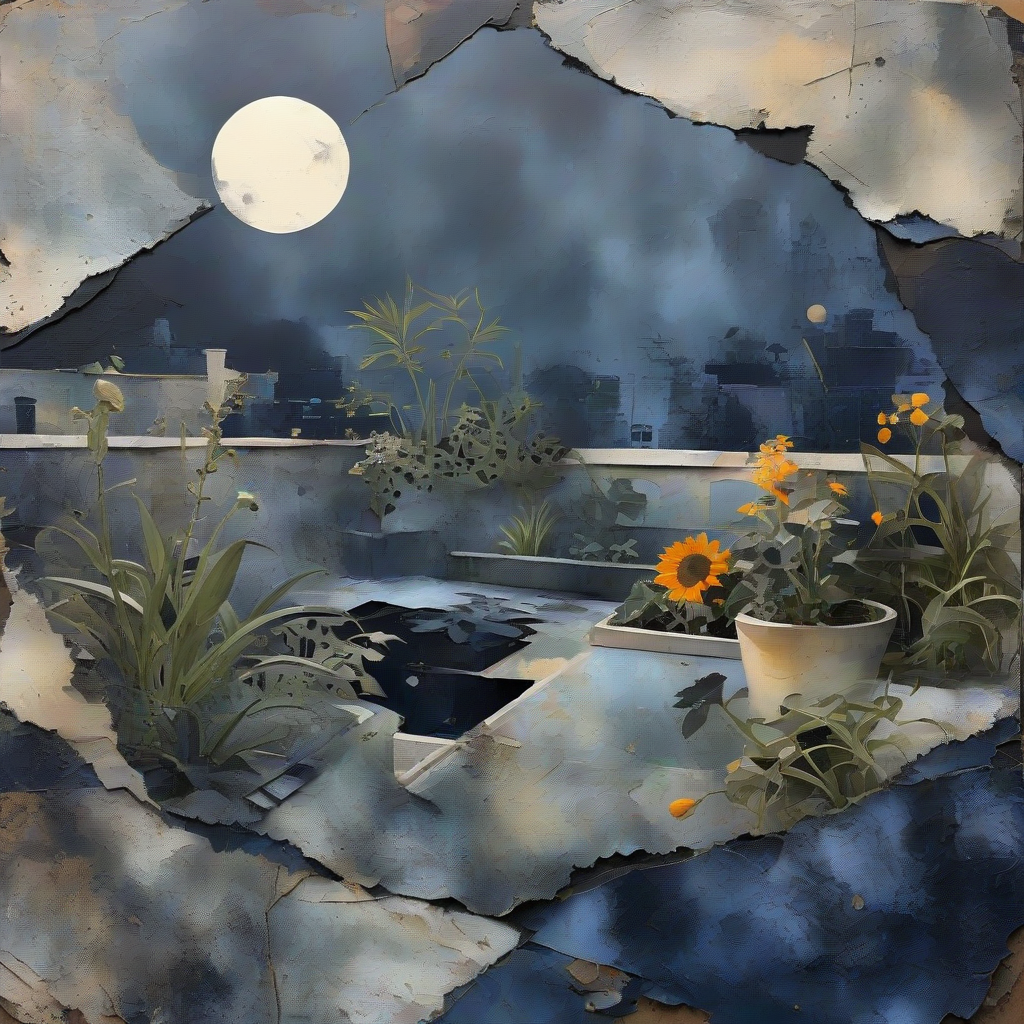}\par
        {\tiny\parbox{\linewidth}{\centering Moonlit rooftop garden with sunflowers, cracked paint, moody blue tones, dreamy atmosphere}}
    \end{subfigure}

    \caption{General artistic style and medium conditioning examples using the same prompt.}
    \label{fig:general-styles}
\end{figure}

\section{Dataset Evaluation}
This section outlines the quantitative evaluation protocols used to characterize the Lunara dataset. We assess four complementary aspects that are critical for vision–language atasets: visual aesthetic quality, image–text semantic alignment, cross-modal retrieval behavior, and visual diversity. Together, these analyses provide a structured framework for examining both the perceptual properties of the images and the semantic fidelity of their associated prompts.

\subsection{Aesthetic Preference Analysis.}
We evaluate image aesthetics using the LAION Aesthetics v2 predictor, a CLIP-based model trained to approximate aggregate human judgments of visual appeal.
We compare our dataset against several widely used vision--language datasets:
Conceptual Captions (CC3M) (\cite{sharma2018conceptual}),
a random subset of LAION-2B-Aesthetic (\cite{schuhmann2022laion}),
and the Wikipedia-based Image--Text dataset (WIT) (\cite{srinivasan2021wit}).
Table~\ref{tab:aesthetic_distribution} reports full distributional statistics of predicted aesthetic scores.
Our dataset (Lunara) achieves a substantially higher mean aesthetic score (6.32) than all baselines,
exceeding CC3M by +1.54, LAION-2B-Aesthetic by +1.07, and WIT by +1.24.
In addition to higher central tendency, Lunara exhibits a markedly shifted distribution:
its median score (6.31) exceeds the 95th percentile of CC3M and closely approaches the upper tail of LAION-2B-Aesthetic and WIT.
Importantly, Lunara contains a significantly larger proportion of highly aesthetic images.
Approximately 33.99\% of images exceed the commonly used threshold of 6.5,
compared to none for CC3M, 0.2\% for LAION-2B-Aesthetic, and 0.1\% for WIT.
This represents over two orders of magnitude enrichment in high-aesthetic content relative to existing large-scale datasets.
Consistent improvements are also observed across the lower tail:
the 5th percentile score of Lunara (5.54) surpasses the median scores of all comparison datasets,
indicating that aesthetic quality is maintained throughout the distribution rather than driven by a small subset of outliers.
Together, these results quantitatively validate the dataset’s design objective:
to prioritize visually rich, high-quality imagery rather than broad, noisy web coverage.
The strong gains across mean, median, tail statistics, and threshold-based metrics
suggest that Lunara offers a substantially cleaner and more aesthetically aligned training signal
for vision--language and generative modeling tasks.

\begin{table}[t]
\centering
\small
\begin{tabular}{lccccccc}
\toprule
\textbf{Dataset} & \textbf{N} & \textbf{Mean} & \textbf{Std} & \textbf{P05} & \textbf{P50} & \textbf{P95} & \textbf{\% $\geq$ 6.5} \\
\midrule
Lunara & 2000 & \textbf{6.32} & 0.49 & \textbf{5.54} & \textbf{6.31} & \textbf{7.18} & \textbf{33.99} \\
CC3M & 1000 & 4.78 & 0.60 & 3.76 & 4.80 & 5.74 & 0.00 \\
LAION-2B-Aesthetic & 1000 & 5.25 & 0.44 & 4.55 & 5.25 & 5.95 & 0.20 \\
WIT & 1000 & 5.08 & 0.57 & 4.13 & 5.09 & 6.04 & 0.10 \\
\bottomrule
\end{tabular}
\caption{
Full distribution statistics of LAION Aesthetics v2 scores.
P05 and P95 denote the 5th and 95th percentiles, respectively.
The final column reports the percentage of images exceeding the commonly used
aesthetic threshold of 6.5.
}
\label{tab:aesthetic_distribution}
\end{table}

\subsection{Image--Text Alignment}
We measure image--text semantic alignment using CLIP cosine similarity (\cite{radford2021clip}). Using a ViT-B/32 backbone pretrained on OpenAI data, the dataset achieves a mean cosine similarity of 0.317 $\pm$ 0.025, indicating consistent alignment between images and their corresponding captions. The relatively low standard deviation suggests that caption quality and semantic grounding are stable across samples and categories.

Per-category analysis shows comparable alignment scores across all categories, with means ranging from 0.33 to 0.38. This consistency indicates that no individual category suffers from systematically weaker captions or mismatched semantics, supporting the dataset’s suitability for controlled evaluation.

While absolute CLIP similarity values are not directly comparable across different CLIP backbones or datasets, prior work has shown that CLIP-based similarity remains a reliable relative indicator of image--text alignment within a fixed evaluation setup (\cite{hessel2021clipscore,ilharco2021clip}). The observed scores are consistent with expectations for stylistic and descriptive captions that emphasize aesthetic attributes (e.g., lighting, artistic style) rather than purely object-centric descriptions.

\subsection{Cross-Modal Retrieval}
To further assess alignment quality, we evaluate bidirectional cross-modal retrieval following standard CLIP-based evaluation protocols (\cite{radford2021clip}). On the same subset, text-to-image retrieval achieves Recall@1 = 43.07\%, Recall@5 = 76.37\%, and Recall@10 = 85.29\%, with a median rank of 2.0. Image-to-text retrieval yields slightly lower but comparable performance, with Recall@1 = 41.87\% and a median rank of 2.0, which is typical for image-to-text retrieval tasks (\cite{karpathy2015deep}).

These results indicate that, in most cases, the correct image--caption pair is retrieved within the top few candidates. The relatively strong Recall@10 values reflect robust semantic grounding, even in the presence of visually similar samples. We note that the dataset contains many portrait-style images with overlapping visual themes, which inherently increases retrieval difficulty and can suppress Recall@1 despite accurate captions (\cite{ilharco2021clip}).

Using a stronger CLIP backbone (ViT-L/14 pretrained on LAION-2B) substantially improves retrieval performance, demonstrating that the dataset benefits from higher-capacity vision--language models and is not bottlenecked by annotation quality (\cite{schuhmann2022laion}).

\subsection{Visual Diversity}
We quantify visual diversity using LPIPS, a learned perceptual similarity metric designed to approximate human judgments of visual similarity (\cite{zhang2018lpips}). The dataset exhibits an average intra-category LPIPS of 0.666 and an inter-category LPIPS of 0.719, computed over randomly sampled image pairs.

The higher inter-category LPIPS confirms that images from different categories are perceptually more distinct than those within the same category, indicating meaningful stylistic and semantic separation. This behaviour is desirable for generative modeling and representation learning, as it suggests that the dataset captures both coherent category-level structure and sufficient visual variability (\cite{zhang2018lpips}).

\begin{table}[t]
\centering
\label{tab:clip_retrieval}
\resizebox{\linewidth}{!}{%
\begin{tabular}{l l c c c c c c c c c}
\toprule
\textbf{Backbone} & \textbf{Pretrain} & \textbf{CLIP sim} & \multicolumn{4}{c}{\textbf{Text$\rightarrow$Image}} & \multicolumn{4}{c}{\textbf{Image$\rightarrow$Text}} \\
\cmidrule(lr){4-7}\cmidrule(lr){8-11}
& & (mean$\pm$std) & R@1 & R@5 & R@10 & MedR & R@1 & R@5 & R@10 & MedR \\
\midrule
ViT-B-32 & openai & 0.319$\pm$0.026 & 43.07 & 76.37 & 85.29 & 2.0 & 41.87 & 74.48 & 85.94 & 2.0 \\
ViT-L-14 & laion2b\_s32b\_b82k & 0.352$\pm$0.041 & 61.47 & 89.93 & 95.76 & 1.0 & 58.18 & 87.79 & 95.31 & 1.0 \\
\bottomrule
\end{tabular}%
}
\caption{CLIP-based alignment and cross-modal retrieval on the dataset (2000 image--prompt pairs). Higher is better for CLIP similarity and Recall@K; lower is better for median rank (MedR).}
\end{table}

\begin{table}[t]
\centering
\label{tab:lpips_diversity}
\begin{tabular}{l c}
\toprule
\textbf{Metric} & \textbf{Value} \\
\midrule
Intra-category LPIPS (mean) & 0.666 \\
Inter-category LPIPS (mean) & 0.719 \\
Pairs sampled (intra / inter) & 2000 / 2000 \\
\bottomrule
\end{tabular}
\caption{LPIPS diversity (AlexNet backbone, images downscaled to 256$\times$256). Higher indicates greater perceptual diversity.}
\end{table}

\section{Discussion}
The Lunara Aesthetic Dataset is designed for prompt grounding and style conditioning research under controlled, reproducible conditions. By pairing model-generated images with human-validated prompts, the dataset enables diagnostic evaluation of prompt adherence and semantic alignment, and supports targeted model adaptation aimed at improving prompt-following behavior.

While this release represents the first publicly available dataset to systematically operationalize regional and cultural art styles for controlled prompt conditioning, artistic traditions are inherently nuanced, historically layered, and internally diverse. As with linguistic variation—where dialects may differ substantially within the same language (\cite{bouamor-etal-2019-madar,abdelali-etal-2021-qadi}) and distinct languages may nonetheless share structural or lexical similarities (\cite{hassan-etal-2022-cross})—artistic variation does not conform to discrete categories but instead exists along a continuous spectrum. Broad regional labels necessarily abstract away important distinctions that exist across time periods, local schools, and socio-cultural contexts. For example, Korean traditional painting differs substantially from Chinese ink traditions in composition, symbolism, and brush technique; similarly, early Indian styles differ markedly from modern Indian art in colour usage, thematic focus, and philosophical grounding. Comparable variation exists across Middle Eastern, Nordic, and Southeast Asian art histories.

Consequently, the styles represented in this dataset are designed to enable controlled experimentation and comparative analysis, not to serve as authoritative taxonomies of regional art. Future releases of the Lunara Art Dataset will expand along temporal, geographical, and stylistic axes, incorporating finer-grained distinctions and historically grounded substyles to better reflect the richness and evolution of artistic traditions.

A key design choice is licensing clarity. Unlike many web-scale resources with mixed or unclear licensing, this dataset is released under Apache 2.0 to encourage broad reuse, including commercial experimentation.

\section{Conclusion and Future Releases}
We presented the Lunara Aesthetic Dataset, a public release of image–prompt pairs curated specifically for the study of aesthetic and style conditioning in text-to-image generation systems. 

Quantitative evaluations demonstrate the dataset has strong aesthetic quality, which serves as a standardized benchmark for assessing aesthetic performance. Evaluations also validated image–text alignment, and controlled stylistic variation. These results confirm the dataset’s suitability for diagnostic evaluation as well as for fine-tuning and adaptation experiments.

To our knowledge, it is the first high-quality open-source dataset to capture nuanced artistic aesthetics spanning cultural regions, historical traditions, and diverse artistic media. This level of curation is particularly important for downstream vision–language learning and model diagnostic analysis, including the evaluation of prompt adherence and fine-grained aesthetic grounding.

% The dataset is released under the Apache 2.0 license to encourage broad academic and commercial reuse. 
Future releases will expand the dataset in semantic variations, scale and stylistic granularity, with deeper coverage of regional, temporal, and medium-specific art traditions.

% The dataset is publicly available at [URL TO BE ADDED].

\section*{Limitations and Ethical Considerations}

The Lunara Art Dataset is intentionally designed to prioritize prompt reliability, and stylistic control over scale and exhaustive coverage. As a result, it should not be interpreted as a comprehensive representation of global artistic traditions or natural language usage.

The dataset consists exclusively of synthetic, model-generated images paired with human-refined prompts. It does not contain depictions of real individuals, biometric identifiers, or personal data, and it is not intended for use in surveillance, biometric recognition, or inference of sensitive attributes.

Because the dataset is curated, it should not be used to draw conclusions about real-world populations, cultures, or artistic communities. While regional styles are included for analytical purposes, they are abstractions and should not be interpreted as definitive or exhaustive representations of any culture or tradition. We encourage users to engage with the dataset critically and responsibly, particularly when conducting analyses related to cultural or aesthetic representation.
\printbibliography

\end{document}